\documentclass[sn-nature]{sn-jnl}% Style for submissions to Nature Portfolio journals
\usepackage{graphicx}%
\usepackage{multirow}%
\usepackage{amsmath,amssymb,amsfonts}%
\usepackage{amsthm}%
\usepackage{mathrsfs}%
\usepackage[title]{appendix}%
\usepackage{xcolor}%
\usepackage{soul} %
\usepackage{textcomp}%
\usepackage{manyfoot}%
\usepackage{booktabs}%
\usepackage{algorithm}%
\usepackage{algorithmicx}%
\usepackage{algpseudocode}%
\usepackage{listings}%
\usepackage{enumitem} %
\usepackage[bbgreekl]{mathbbol} %
\usepackage{amsmath} %
\usepackage{bbm}
\usepackage{comment}
\usepackage{titlesec}
\usepackage[section]{placeins}
\theoremstyle{thmstyleone}%
\theoremstyle{thmstyletwo}%

\theoremstyle{thmstylethree}%

\begin{document}

\title[Article Title]{Many happy returns: machine learning to support platelet issuing and waste reduction in hospital blood banks}


\author*[1]{\fnm{Joseph} \sur{Farrington}}\email{ucabjmf@ucl.ac.uk}

\author[2]{\fnm{Samah} \sur{Alimam}}%\email{iiauthor@gmail.com}

\author[3]{\fnm{Martin} \sur{Utley}}%\email{iiiauthor@gmail.com}

\author*[1]{\fnm{Kezhi} \sur{Li}}\email{ken.li@ucl.ac.uk} \equalcont %\email{iiiauthor@gmail.com}

\author[1,4,5]{\fnm{Wai Keong} \sur{Wong}} \equalcont %\email{iiiauthor@gmail.com}

\affil[1]{\orgdiv{Institute of Health Informatics}, \orgname{University College London}, \orgaddress{\street{222 Euston Road}, \city{London}, \postcode{NW1 2DA}, \country{UK}}}

\affil[2]{\orgdiv{Department of  Haematology}, \orgname{University College London Hospitals NHS Foundation Trust}, \orgaddress{\street{250 Euston Road}, \city{London}, \postcode{NW1 2PG}, \country{UK}}}

\affil[3]{\orgdiv{Clinical Operational Research Unit}, \orgname{University College London}, \orgaddress{\street{4 Taviton Street}, \city{London}, \postcode{WC1H 0BT}, \country{UK}}}

\affil[4]{\orgdiv{NIHR University College London Hospitals Biomedical Research Centre}, \orgname{University College London and University College London Hospitals NHS Foundation Trust}, \orgaddress{\street{149 Tottenham Court Road}, \city{London}, \postcode{W1T 7DN}, \country{UK}}}

\affil[5]{\orgdiv{Clinical Director of Digital Transformation}, \orgname{Cambridge University Hospitals NHS Foundation Trust}, \orgaddress{\street{Hills Road}, \city{Cambridge}, \postcode{CB2 0QQ}, \country{UK}}}


\abstract{Efforts to reduce platelet wastage in hospital blood banks have focused on ordering policies, but the predominant practice of issuing the oldest unit first may not be optimal when some units are returned unused. We propose a novel, machine learning (ML)-guided issuing policy to increase the likelihood of returned units being reissued before expiration. Our ML model trained to predict returns on 17,297 requests for platelets gave AUROC 0.74 on 9,353 held-out requests. Prior to ML model development we built a simulation of the blood bank operation that incorporated returns to understand the scale of benefits of such a model. Using our trained model in the simulation gave an estimated reduction in wastage of 14\%. Our partner hospital is considering adopting our approach, which would be particularly beneficial for hospitals with higher return rates and where units have a shorter remaining useful life on arrival.}

\maketitle

\section{Introduction} \label{sec:intro}

The inventory management of platelets in hospital blood banks is complicated by their short shelf life of 5--7 days after donation \cite{flint_is_2020}, which can lead to high levels of wastage. A recent systematic review found that the common range for wastage rates in hospitals was 10-20\% \cite{flint_is_2020}. The UK (United Kingdom) Blood Stocks Management Scheme reported that 4.5\% of platelet units issued to hospitals in England were wasted at a cost of £2m in 2017/2018 \cite{blood_stocks_management_scheme_nhsbt_2018}, and a higher wastage rate of 4.8\% (11,758 units reported as wasted, of 247,416 units issued) in 2022/2023 \cite{blood_stocks_management_scheme_2023}. In addition to the financial cost of wastage, it can be argued that there is a moral duty to make the best possible use of blood that has been altruistically donated \cite{Yates2017, Stanger2012, tissot_ethics_2016}. In countries with centralised systems such as the UK, hospital blood banks order from a provider that handles the collection of donations and subsequent manufacturing into products that will be transfused to patients. In these systems, hospital blood banks must make two main decisions: replenishment (how many units to order from the central provider) and issuing (which unit to provide for a patient in response to a request from a clinician). They aim to avoid the potential clinical harm of shortages while minimising wastage. The rich, real-time data now available in many hospitals due to the deployment of electronic health record (EHR) systems facilitates the use of machine learning (ML) methods to support these decisions, helping to reduce wastage and ensure that the right unit is available for the right patient at the right time. In this study we focus on using machine learning (ML) to support issuing decisions that account for possibility of returns: a flow of stock back into the hospital blood bank when requested units are issued but not transfused.

Dumkreiger \cite{Dumkreiger2020} highlighted the lack of research on issuing policies compared to replenishment policies in the blood product inventory literature. Most studies use an Oldest-Unit First-Out (OUFO) issuing policy \cite{Osorio2015, Piraban2019, torrado_towards_2022}  which is intuitively optimal \cite{haijema_optimal_2014} and recommended best practice for blood products \cite{stanger_what_2012}. This observation is true for the wider perishable inventory literature, in which most studies use a OUFO issuing policy if, as in a hospital blood bank, the products are selected  by the supplier \cite{bakker_review_2012, chaudhary_state---art_2018, janssen_literature_2016, karaesmen_managing_2011, mirabelli_optimization_2022, Nahmias1982}. More complex policies have been utilised in studies considering a preference for fresher units for a subset of patients \cite{pierskalla_optimal_1972, dalalah_platelets_2019, civelek_blood_2015}, the potential for substitution between blood groups if a direct match is not available \cite{Dillon2017, duan_optimization_2014, meneses_blood_2021}, uncontrolled replenishment \cite{abbasi_issuing_2014} and reducing the mean age of transfused blood \cite{atkinson_novel_2012}. Dumkreiger \cite{Dumkreiger2020} and Abdulwahab and Wahab \cite{Abdulwahab2014} used approximate dynamic programming to learn custom issuing policies for packed red blood cells (PRBC) and platelets respectively based on the patient's blood group and current stock holding, but neither considered returns or the likelihood of use. An OUFO policy may not be optimal when some units are returned: an older unit issued to a patient less likely to be transfused may expire before it can be used. In preliminary work for this study, we found that 8\% of platelet units issued in 2015 and 2016 by the hospital blood bank at our partner site, University College London NHS Foundation Trust (UCLH), were not transfused.

The challenge of platelet returns has not been considered in the literature to our knowledge and was absent from a recent systematic review of efforts to reduce platelet wastage \cite{flint_is_2020}. Prior to the development of electronic cross-matching and remote issuing \cite{staves_electronic_2008}, studies often included the related concept of assigning PRBC units to a sub-inventory for a specific patient following a physical cross-match, with units returned to the main inventory if not used within a specified period  \cite{jennings_analysis_1968, pereira_blood_2005, rytila_using_2006, Katsaliaki2007}. Previous studies have also considered  the return of units to a regional blood centre for reissue to a hospital with higher demand \cite{yahnke_analysis_1972, yahnke_analysis_1973, britten_weekly_1979, chen_managing_2022, ahmadimanesh_designing_2022}, but this practice is not permitted in many high-income countries \cite{Osorio2015} and is not considered here. Product returns have been extensively studied in other contexts \cite{ambilkar_product_2021, abdulla_taking_2019}, but forecasting efforts have focused mainly on aggregate return volumes \cite{canda_modeling_2015, tsiliyannis_markov_2018, chou_policies_2020, cui_predicting_2020} and recent work predicting returns at the individual level for e-commerce clothing retailers \cite{Joshi2019, zhu_local_2018, li_trust-aware_2019, fu_fused_2016} was reasonably not used to inform issuing decisions. 

We propose a novel, ML-guided issuing policy that incorporates the possibility of returns: issue the youngest unit if a model predicts that the request will not result in transfusion, and the oldest unit otherwise. With a sufficiently good predictive model this policy should increase the likelihood that, if a unit is returned, it can be reissued to another patient before it expires. The closest related work we have identified is a working paper by Brodheim and Prastacos \cite{brodheim_demand_1980}, who proposed a method to assign older PRBC units to sub-inventories for patients who were most likely to receive a transfusion, based on the quantity requested, at a time when physical cross-matching was required. We focus on platelets, and the development of an ML model trained on a wide range of features to predict whether requested units will be transfused. 

Shah et al. \cite{shah_making_2019} observed that conventional model evaluation, using performance metrics such as sensitivity, specificity and the area under the receiver operating characteristic curve (AUROC), measure the quality of an ML model's predictions but provide limited information about the impact of those predictions on patient care and costs. ML models achieving high scores on performance metrics have been found to achieve limited or no clinical benefit in randomised controlled trials \cite{zhou_clinical_2021}. It is therefore critical to consider the clinical workflow in which the ML model would be deployed, including changes to current practice which may be enabled by the predictions \cite{li_developing_2020}.  Recent work has addressed this challenge by incorporating trained models into a simulated workflow to estimate the potential impact of the model in terms of key performance indicators (KPIs) of interest to decision makers \cite{misic_simulation-based_2021,wornow_aplus_2023}, but significant work is often required to get access to the data required in order to begin the ML model development process in healthcare \cite{taylor_road_2021}. We therefore adopted a simulation-first approach: simulating the predictions of hypothetical ML models in a simulated workflow to understand the effect of different levels of predictive model performance on KPIs, and to determine if even a perfect predictive model would have appreciable beneficial impact. This could be considered a method to estimate the value of information provided by a prediction with a specified level of quality, by comparing the KPIs of the system using the prediction to those without the prediction \cite{viet_value_2018}.

Simulation has been widely used within the blood product inventory management literature to evaluate benefits of changes in workflow or procedures \cite{rytila_using_2006, Katsaliaki2007, Kopach2008, yuzgec_simulation_2013, katz_simulation_1983, asllani_simulation-based_2014, baesler_analysis_2014}, different replenishment policies \cite{rajendran_hybrid_2020, duan_new_2013, blake_optimizing_2003, haijema_blood_2007} and, recently, the potential benefits of demand forecasting models using real-time data from EHRs \cite{li_decision_2021, motamedi_demand_2021, schilling_reduction_2022, Guan2017d, abouee_mehrizi_data-driven_2022} in terms of KPIs \textit{after} models have been trained.  Within the broader supply chain literature, researchers have investigated the effect of different levels of error in a demand forecast on KPIs \cite{fildes_incorporating_2011, altendorfer_effects_2016, sanders_quantifying_2009}. Dumkreiger \cite{Dumkreiger2020} simulated PRBC demand forecasting models with different specified levels of error to determine whether better forecasts could reduce wastage and shortages - the only work we have identified investigating the potential benefits of a predictive model to blood product management prior to model development. 

This paper describes how we trained and evaluated a first ML model to predict platelet returns, having first investigated the potential operational utility of such a model with different levels of predictive performance using a simulation-first approach. In addition to these findings, the contributions of the research include the development of a model of a platelet inventory management workflow that includes returns (an important aspect of the real problem previously neglected in the literature). The key inputs to the simulated workflow are based on observed data from the hospital blood bank at our partner hospital and data previously reported in the literature.  Our partner hospital is considering adopting a version of our proposed issuing policy based on these results. 

\section{Results} \label{sec:results}

Our simulation experiments indicated that a predictive model for platelet returns could reduce wastage when used to inform decisions taken with our proposed Youngest Unit for Predicted Returns (YUPR) issuing policy (see Supplementary Note \ref{app:sim_exp_results_all}). We therefore proceeded to develop the predictive model and to evaluate its potential impact in terms of  KPIs. 

\subsection{A predictive model for platelet returns} \label{sec:results:pred}

\begin{figure}[h] 
\centering
\includegraphics[width=1.0\textwidth]{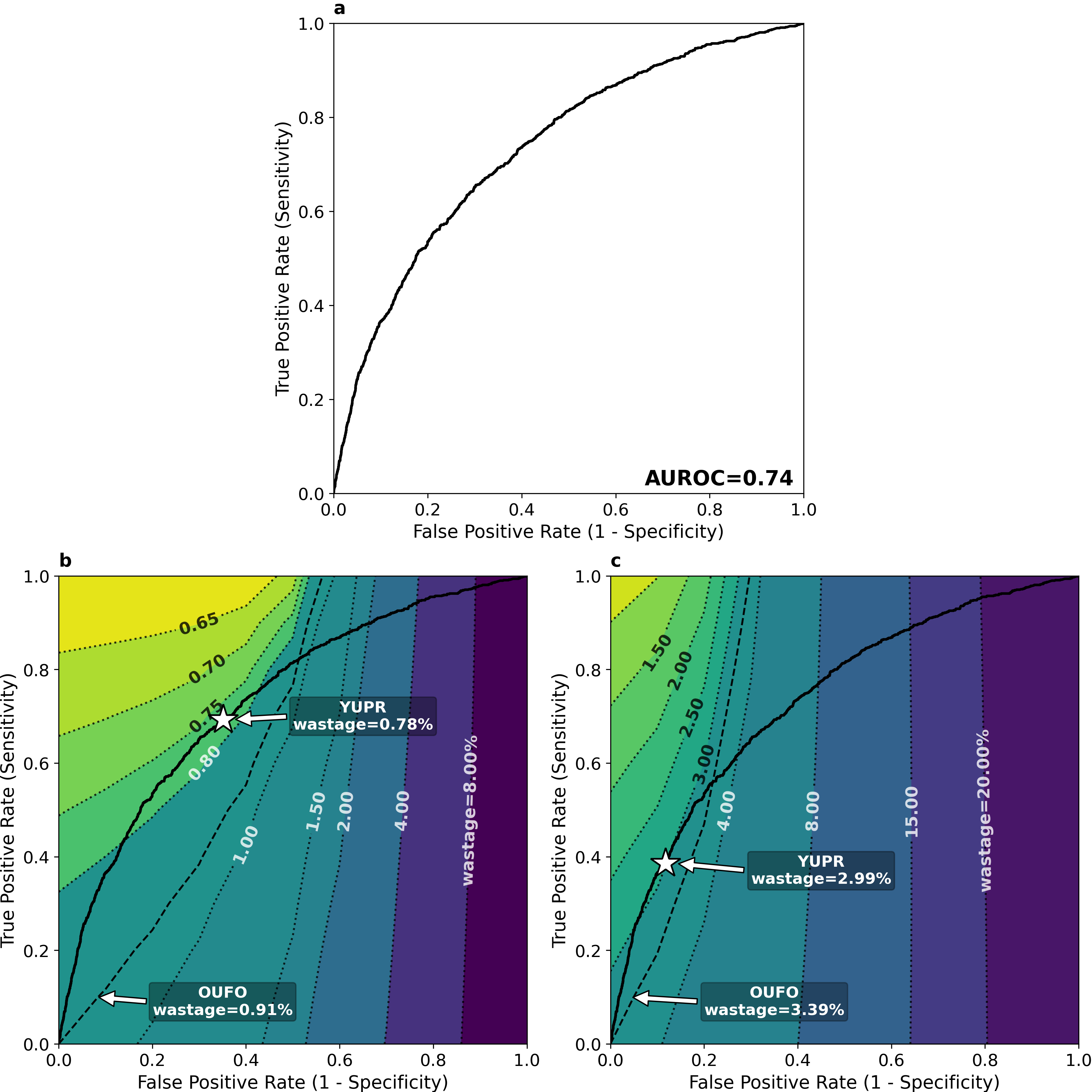}
\caption{\textbf{The ROC curve of the trained predictive model on the test set, platelet requests for units required in 2017, and contour plots illustrating estimated wastage when issuing platelets using the trained predictive model within our proposed YUPR issuing policy}. The receiver operating characteristic (ROC) curve on the test set plotted alone \textbf{(a)}, and overlaid on a contour plot of wastage generated by our simulation of the hospital blood bank workflow assuming \textbf{(b)} the distribution of remaining useful life on arrival observed at UCLH and \textbf{(c)} the distribution of remaining useful life on arrival reported for a US hospital by Rajendran and Ravindran \cite{rajendran_platelet_2017}. Under both settings, the predictive model could achieve lower wastage than an OUFO policy with no reduction in service level. A larger absolute reduction in wastage is possible in \textbf{(c)}, in which wastage is higher due to the shorter average remaining useful life on arrival. Lighter colours indicate better performance. The region above and to the left of the contour for an OUFO issuing policy comprises combinations of sensitivity and specificity required for the YUPR issuing policy to incur lower wastage that OUFO. The ROC curve represents possible combinations of sensitivity and sensitivity that could be achieved by the selection of different thresholds for distinguishing positive and negative predictions, and the white star indicates the predictive performance that would be achieved on the test set when selecting a threshold to minimise wastage using the training set ROC curve.} \label{fig:contour_test}
\end{figure}

Trained on 17,297 requests for platelets required between 1 February 2015 and 31 December 2016, our ML model for predicting whether at least one unit would go unused in each request gave an AUROC of 0.74 in the test set of 9,353 requests for platelets required in the calendar year 2017 (Figure \ref{fig:contour_test}a). 

\begin{figure}[h!] 
\centering
\includegraphics[width=1.0\textwidth]{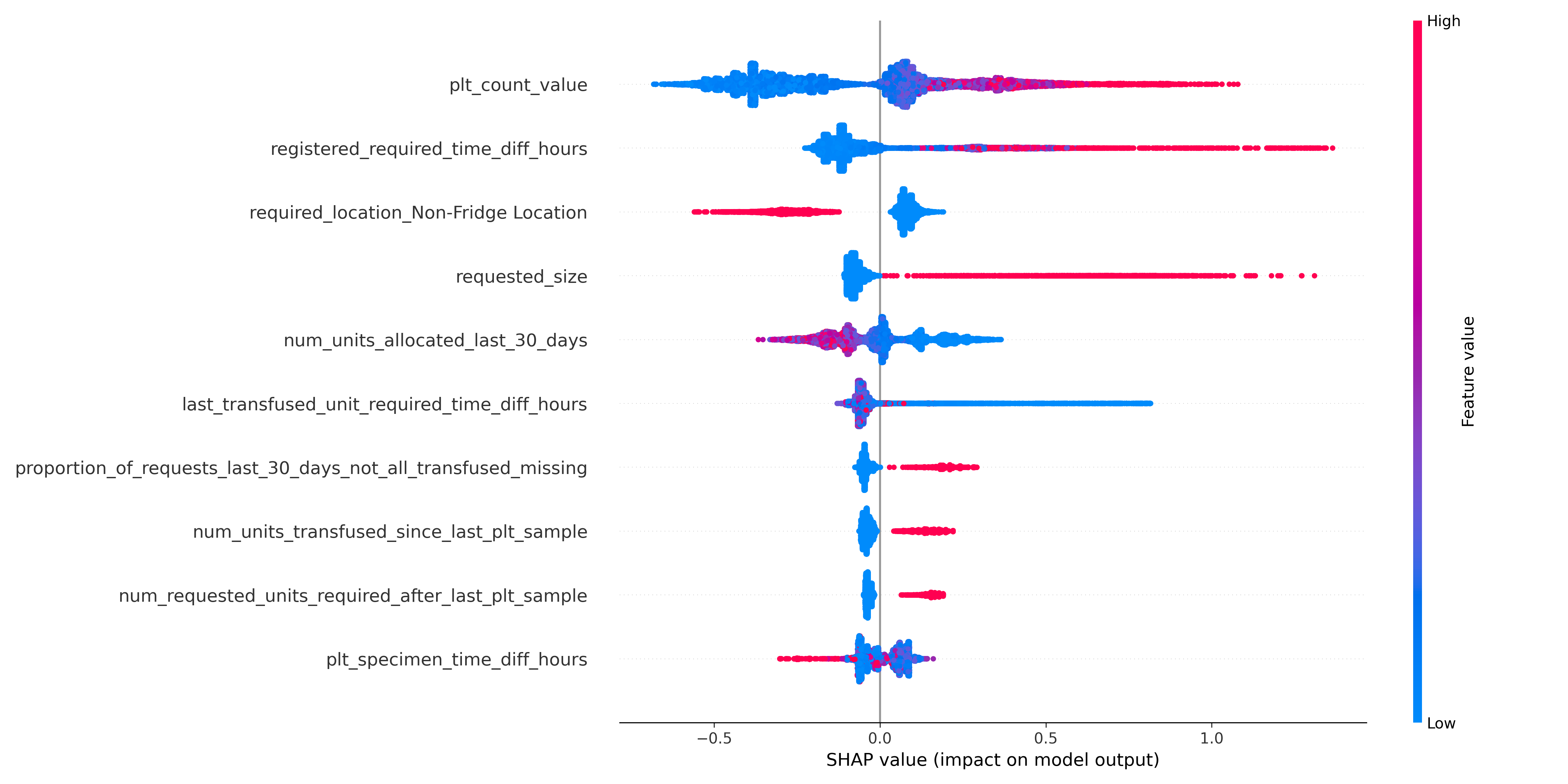}
\caption{\textbf{Summary plot of SHAP values for the 10 most important features in the trained prediction model.} Predictions were made for each request, with patient-level features, such as ``plt\_count\_value'' (the patient's platelet count) based on the latest available data for the patient at the time the issuing decision is made. The feature importance values were computed using the training set. Each point represents an example from the training set. A large positive SHAP value means that the feature pushed the model output towards a positive prediction. A description of each feature is set out in Table \ref{tab:feature_summary} in Supplementary Note \ref{app:ml_features}.} \label{fig:shap}
\end{figure}

Figure \ref{fig:shap} shows the Shapley Additive Explanations (SHAP) \cite{Lundberg2017a} values of feature importance. The most important feature in the model was the value of the most recent platelet count, with higher platelet counts associated with predicted returns. Seven of the top 10 features were hand-engineered features, including the second most important feature of how long in advance a request was made.

\subsection{Impact of an ML-guided issuing policy on wastage and service level}

By simulating the workflow of a hospital blood bank as it orders and issues platelets in response to requests from clinical teams, we estimated that use of the predictive model would reduce wastage by 14\% when using simulation inputs derived from our partner hospital (from 0.91\% to 0.78\%) with no detriment to service level (Figure \ref{fig:contour_test}b).

If the useful life of platelets on arrival to the blood bank were as reported in a US hospital \cite{rajendran_platelet_2017}, use of the model would reduce wastage by 12\% with no detriment to service level (Figure \ref{fig:contour_test}c). This reduction is from a higher starting value of wastage, 3.39\% under an OUFO issuing policy, due to the shorter average remaining useful life on arrival.

The contour plots superposed onto the ROC curve in Figures \ref{fig:contour_test}b and \ref{fig:contour_test}c show how future improvements in predicting returns could give sizeable reductions in wastage. 

The potential for wastage reduction using the trained predictive model was confirmed by supporting analysis incorporating real demand, and the model predictions for the associated requests, for the calendar year 2017 into the simulated workflow (see Supplementary Note \ref{app:real_demand_eval}).

In this work we used a replenishment policy where orders are based on the number of units in stock, rather than one in which fixed number of platelet units were ordered every day, because this policy gave lower wastage and a higher service level in our initial simulation experiments investigating the interaction between the replenishment and issuing policies (see Table \ref{tab:basic_sim_pairwise} in Supplementary Note \ref{app:sim_exp_results_all}). 

Sensitivity analyses shows that our predictive modelling and YUPR issuing policy may be particularly effective at reducing wastage in situations where a higher proportion of requested units are not transfused and when the remaining useful life of platelet units on arrival is shorter on average (see Figures \ref{fig:scenario_plot} and \ref{fig:pairwise_scenario_plot} in Supplementary Note \ref{app:sim_exp_results_all}).

 \section{Discussion} \label{sec:discussion}

To our knowledge this work presents the first patient-level predictive model of whether platelets issued in response to a request are actually transfused. Indeed, we believe this to be the first paper to even consider the fact of post-issue return of platelets within hospital blood banks and consider how this fact may change the most appropriate issuing policy. Our results show that an ML-guided issuing policy could reduce wastage when it is possible to reissue returned units. The projected benefit of a 14\% reduction in wastage for our partner hospital demonstrates that the approach would be beneficial even in the context of the low wastage they currently achieve, and our sensitivity analyses highlight that when units have a shorter remaining useful life on arrival or when the return rate is higher, our approach can lead to large absolute reductions in wastage. Our YUPR issuing policy offers wastage reductions of up to 80\% (with a perfect predictive model) for hospital receiving platelets with a remaining useful life on arrival reported in a US hospital \cite{rajendran_platelet_2017}. Under the proposed issuing policy the KPIs are similar for both the US and locally observed distributions of remaining useful life on arrival despite the much shorter useful life on arrival in the US case (see Table \ref{tab:basic_sim_pairwise} in Supplementary Note \ref{app:sim_exp_results_all}). The UCLH transfusion laboratory is considering moving to a YUPR policy, although they are exploring a rule-based prediction model based on augmenting the insights from our ML model with expert knowledge rather than investing at this point in integrating different software systems to support real-time ML model predictions. 

Our results demonstrate the importance of considering returns when evaluating policies for managing platelet inventory. Policies based solely on the number of transfused units could lead to shortages because they underestimate the actual demand when not all requested units are used. Consideration of returns, and our proposed issuing policy, may be less impactful at sites where issued units that are not transfused spend a shorter time away from the decision-making point. It may have a greater impact as improvements in technology or procedures increase the proportion of returned units that can be reissued \cite{kron_multicenter_2021}. 

This study also provides a compelling example of using a simulation-first approach to evaluate the extent to which an ML model can support a clinical or operational workflow. The simulated workflow was used in three ways: firstly to estimate the potential benefits of a proposed predictive model in the workflow, secondly to inform model selection during hyperparameter search and thirdly to estimate the performance of a trained model on a real time period using established KPIs. The results from the simulation informed the subsequent steps of our research by establishing the potential gains on offer and establishing the performance required for our proof-of-concept ML model to be beneficial. The results highlight the importance of investigating how a predictive model performs in a workflow in terms of KPIs as well as evaluation using ML metrics. The same predictive model, with the same AUROC, can have a bigger impact on the workflow under some circumstances, such as under the distribution of remaining useful life on arrival observed by Rajendran and Ravindran \cite{rajendran_platelet_2017}, than others. Our work builds on recent work using simulation to evaluate trained ML models in terms of KPIs that have real-world impact rather than ML metrics \cite{wornow_aplus_2023, misic_simulation-based_2021} but with the added potential to understand whether the predictions could be useful \textit{before} building the predictive model and to understand how well a predictive model needs to perform to achieve a specified improvement in one or more KPIs. A simulation-first approach therefore offers a route for early evaluation of predictive model utility, helping prioritize data access requests and development efforts for models likely to improve KPIs when integrated into clinical or operational workflows. This is analogous to using estimates of the value of information to target clinical research in health economics \cite{fenwick_value_2020}. A further advantage of a simulation-first approach is that the estimated KPIs can help to focus the hyperparameter search on models that are likely to have a higher utility. While our workflow requires simulation, for other some workflows, such as  those that can be modelled as a simple queue, the impact of a predictive model of a specified quality on KPIs can be investigated analytically \cite{feizi_vertical_2023}. A focus on evaluating potential utility may help to bridge the gap between the large number of ML models that are developed and the relatively small number that have been successfully deployed \cite{seneviratne_bridging_2020}.

This work has a number of limitations. Firstly, we only considered a single type of platelet and assumed that any unit could fill the demand from any patient. This is a common assumption in the literature \cite{Guan2017d, abouee_mehrizi_data-driven_2022, rajendran_platelet_2017}, but neglects compatibility between the blood type of the donor and the patient and the fact that some patients have special transfusion requirements. Our YUPR issuing policy could be adapted to incorporate blood types by, for example, issuing the youngest matching unit or youngest compatible unit for predicted returns. The proposed issuing policy may be less effective when including patients with special requirements because there is less (or no) choice about which unit to issue. We assumed there is no medical reason to allocate fresher platelets to specific patients. Previous studies have considered age-differentiated demand \cite{dalalah_platelets_2019, civelek_blood_2015} with fresher units preferred for patients with certain conditions, but a recent systematic review found no relationship between platelet storage time and clinical outcomes in critically ill or haematology patients \cite{aubron_platelet_2018} and this distinction is not made in UK guidance \cite{estcourt_guidelines_2017, national_institute_for_health_and_care_excellence_blood_2015}. Our proposed issuing policy relies on the fact that issued units are kept in conditions that generally allow them to be reissued if they are not transfused and have not expired. This assumption is valid for our partner hospital UCLH (evidenced by the low estimated slippage rate, see Supplementary Note \ref{app:sim_inputs}), potentially due to the use of remote platelet agitators located around the hospital, but this may not be applicable at all sites. 

We have assumed that the hospital blood bank staff would always follow the replenishment and issuing policies, but Wornow et al. \cite{wornow_aplus_2023} noted that predictions may not be acted upon in practice. Presenting the possible benefits in terms of meaningful and familiar KPIs instead of ML performance metrics could help to build trust in a policy based on predictive models and increase adherence to the policy. 

In evaluating different issuing policies, we used simple heuristic replenishment policies that do not account for the age profile of the stock or any knowledge about the number and age profile of units issued during the day that may be returned. We note that it may be possible to achieve better performance with an OUFO issuing policy using a more sophisticated replenishment policy that takes these factors into account, or by using the return predictions to support both issuing and replenishment decisions. 

We focus our discussion of performance on KPIs of wastage and service level because they are easily interpretable in a way that the notional costs assigned to wastage, shortages and holding are not. As with other literature in this area, the relative costs assigned to these different inventory events do not have an empirical underpinning. The relative costs partly reflect the decision makers' concern about consequences with intangible costs that are difficult to determine, such as the impact on a patent's health caused by a a delay in treatment due to a shortage or loss in donor confidence due to high wastage \cite{Blake2010}. 

The absolute levels of wastage in our simulated system are low relative to previously reported figures but, when using the OUFO issuing policy, are similar to those observed at UCLH during the time period under consideration. Additionally, we optimised the replenishment policies within our simulated workflow and so, given that over-ordering is a commonly cited reason for wastage \cite{sekhar_effective_2016, abbaspour_simple_2021}, may have underestimated the benefits of our approach. 

Given the potential for our proposed policy to mitigate the wastage associated with receiving older stock, future work could investigate whether it is possible to achieve system-wide improvements by directing older units to sites able to implement such a policy. Future work considering ML-based support for platelet issuing could focus on improving the predictive model. One possibility is the inclusion of additional features representing diagnoses, procedures, and the indication for the planned transfusion, where they are available at the time the issuing decision is made. A similar predictive model could also be used to support the decision about whether to issue a platelet unit in response to a request, not just which unit to issue, supporting the work of a platelet coordinator \cite{sekhar_effective_2016} to ensure that blood products are used appropriately and to reduce wastage. This could reduce wastage even in settings where units are not generally able to be reissued but would bring the use of predictive models closer to decisions on the clinical management of individual patients. Such use would, appropriately, come with additional ethical, professional and regulatory considerations.

In conclusion, we developed a first ML predictive model for platelet returns and incorporated it into a novel issuing policy. By modelling the platelet inventory management workflow in a hospital, incorporating the observed flow of some issued units being returned unused, we demonstrated that the predictive model is sufficiently good to reduce wastage, while maintaining service level, when used to inform issuing decisions under our proposed policy. A policy of issuing youngest units for predicted returns could therefore support efforts to reduce platelet waste while the broader concept of a simulation-first approach may help to target future development of predictive models to applications most likely to be beneficial in practice.

\section{Methods} \label{sec:methods}

\subsection{Setting}

The data used in this study comes from our partner hospital trust, UCLH, and its associated transfusion laboratory. The transfusion laboratory is run by Health Services Laboratories and serves the six UCLH hospitals in central London. The transfusion laboratory is a hospital blood bank responsible for ordering, storing, processing, managing and issuing fresh blood products. It also performs blood group, antibody and antigen testing. We describe the inventory management practices at UCLH in Supplementary Note \ref{app:uclh_processes}.

Data was extracted for the period 2015 to 2017 inclusive, from Bank Manager (the transfusion laboratory information system) and the UCLH Archive Data Store. During this period, 1.6\% of the platelet units received by UCLH were wasted. This figure is much lower than both the 10\--20\% reported by Flint et al. \cite{flint_is_2020}, and the 4.5\% reported by the UK Blood Stocks Management Scheme for 2017/2018 \cite{blood_stocks_management_scheme_nhsbt_2018}.

The problem we explore in this work, the return of issued platelet units, was inspired by our conversations with staff at UCLH and the transfusion laboratory. We used data from UCLH and the transfusion laboratory to estimate realistic values of key input parameters. As a first effort to incorporate the concept of returns, our simulated workflow builds on previous platelet inventory management simulations \cite{rajendran_hybrid_2020, mirjalili_2022_data-driven_paper}. The simulated workflow is described in Section \ref{sec:results:workflow}. This workflow was used to conduct initial simulation experiments (see Section \ref{methods:sim:sim_exp}), the results of which informed hyperparameter and classification threshold selection for the predictive model, and to evaluate the performance of the trained predictive model in terms of KPIs.  

\subsection{Developing a predictive model} \label{subsec:methods:pred}

\subsubsection{Data}

We developed a binary classification model to predict, for each request, whether at least one requested unit would be returned. Demographic features (sex, year of birth) and platelet count test results were extracted from the UCLH Archive Data Store. Data related to requests for platelets, and individual platelet units, was extracted from the UCLH transfusion laboratory information system Bank Manager. Missing values for numeric values were imputed using the median and missing values for binary features were imputed using the mode. Binary missing indicator features were added corresponding to each numeric or binary feature. Categorical features were one-hot encoded, and the number of one-hot features for high-cardinality features (defined as 16 or more categories) was determined as part of the hyperparameter tuning process. 

The training set consisted of 17,297 requests with a required date  between 1 February 2015 and 31 December 2016, while the test set consisted of 9,353 requests with a required date between 1 January 2017 and 31 December 2017. We excluded requests where no unit was assigned, where a neonatal unit was assigned, where the patient identifier was that of a test patient used for internal system checks and where the request was required more than 30 minutes before it was registered. Some calculated features are based on a look-back period of 30 days and therefore we also excluded the requests from January 2015 to ensure a full look-back period for all requests in the training and test sets. The time point of the prediction, used to determine the most recent platelet count and what other information was available when the prediction was made, was set as the later of one hour before the request was required, and the time when the request was registered in Bank Manager. See Table \ref{tab:ml_exclusion} in Supplementary Note \ref{app:ml_exclusion} for a breakdown of exclusions, Table \ref{tab:feature_summary} in Supplementary Note \ref{app:ml_features} for a description of all of the input features and Tables \ref{tab:num_bin_features} and \ref{tab:cat_features} in Supplementary Note \ref{app:ml_features} for details of the features including the percentage of missing entries, the mean and standard deviation of numeric features, and the most common categories for categorical features.  

\subsubsection{Training and evaluation}

We used XGBoost \cite{Chen2016}, a popular and fast implementation of gradient boosting with decision trees which is known to perform well on a variety of prediction tasks based on tabular data \cite{shwartz-ziv_tabular_2022}. The hyperparameters were tuned using 10-fold cross-validation over the training set using Optuna's \cite{akiba_optuna_2019} Tree-structured Parzen Estimator (TPE) sampler, with 200 trials. The folds were stratified so that each fold received approximately the same ratio of positive to negative examples, and each patient was assigned to only one fold. Based on the results of the initial simulation experiments (see Section \ref{methods:sim:sim_exp} and Supplementary Note \ref{app:sim_exp_results_all}), we knew that the false positive rate of the model needed to be less than 0.6 for an issuing policy supported by the model to achieve lower daily cost and wastage than an OUFO issuing policy. We therefore used the partial-AUROC \cite{ma_use_2013} to compare model performance during hyperparameter tuning, considering only the area under the curve where the false positive rate was less than 0.6. An improvement in this metric should lead to an improvement in our KPIs whereas an improvement in standard AUROC could be due to better predictive performance in a region of the of the curve that the simulation results suggest will not lead to better inventory management performance. The hyperparameter search ranges, and final selected hyperparameters are set out in Table \ref{tab:ml_hps} in Supplementary Note \ref{app:ml_hps}.  

The final model was trained on the entire training set using the combination of hyperparameters that achieved the highest mean partial-AUROC over the 10 cross-validation folds, and applied to predict outcomes for the test set. Performance on the test set was reported in terms of AUROC because we consider this metric both more familiar and easier to interpret than partial-AUROC.

One approach to estimating the potential benefit of the trained model in terms of KPIs is to calculate the sensitivity and specificity on the test set and look up the KPIs for a simulated predictive model with those metrics using the results from the initial simulation experiments. The sensitivity and specificity depend on the threshold used to determine a positive prediction, and the ROC represents the different possible combinations of sensitivity and specificity as the threshold is increased. The wastage reductions reported in Section \ref{sec:results} are at classification thresholds selected to minimise the estimated wastage using the predictions on the training set. For each potential model threshold in the training set, we estimated the expected wastage by linearly interpolating between the estimated wastage for fixed combinations of sensitivity and specificity and selected the threshold with the minimum estimated wastage. We selected a threshold to minimise wastage directly because the contour plots of service level from the initial simulation experiments showed that the sensitivity and specificity of the predictive model had minimal impact on service level (see Figure \ref{fig:sim_contour} in Supplementary Note \ref{app:sim_exp_results_all}). The sensitivity and specificity on the test set were calculated using the identified threshold, and the corresponding wastage was estimated using linear interpolation on the results from the initial simulation experiments. This analysis was performed separately for each of the two distributions of remaining useful life on arrival. Figures \ref{fig:threshold_uclh} and \ref{fig:threshold_rs} in Supplementary Note \ref{app:ml_hps} show the ROC curve for the training set plotted over a contour plot of wastage and estimated wastage plotted directly against the model threshold for the two different distributions of remaining useful life on arrival.

The KPIs observed at a certain level of sensitivity and specificity assumed in the simulation experiments may not correspond to those that would be achieved at the same level of sensitivity and specificity for our trained model because we made the simplifying assumption that the total demand was generated by requests for single units in the initial simulation experiments. In reality, during 2015 and 2016, 7\% of requests were for more than one unit, the majority of these were for two units and the largest request was for four units. When we trained our ML model on real requests we assigned labels based on whether or not all of the requested units were transfused. As a result, there is a discrepancy between sensitivity and specificity as defined for the simulation and for the real predictive model: for example, a correct positive prediction on a real request for two units could mean that one is transfused and one is returned. We addressed this by performing an additional evaluation step: applying the predictive model to real demand data from 2017 within the simulated workflow, making predictions at the request level (so, for example, a request for two units with a positive prediction would be met by issuing the two freshest units), and calculating the KPIs directly.

We created a version of the simulated workflow (see Section \ref{sec:results:workflow}) that used the real observed demand and returns from 2017 (instead of sampling them), and could use predictions from our trained model to support the issuing decision. Morning demand consisted of requests that were required after 00:00 and before 12:00, and afternoon demand consisted of requests required after 12:00 and before 00:00. The slippage rate (see Section \ref{sec:methods:sim}) was set to $\phi=7\%$ as in the initial simulation experiments. For each of the two distributions of remaining useful life on arrival used in the initial simulation experiments (see Section \ref{methods:sim:sim_exp}) we compared the performance of our YUPR issuing policy, informed by the predictions of the trained predictive model, with an OUFO issuing policy. The replenishment policy was an \texttt{(s,S)} replenishment policy (see Section \ref{sec:rep_policy}) with parameters fit during the initial simulation experiments. The classification thresholds for the predictive model were determined as set out above. 

\subsection{Simulation}\label{sec:methods:sim}

\begin{figure}[h]
\centering
\includegraphics[width=1.0\textwidth]{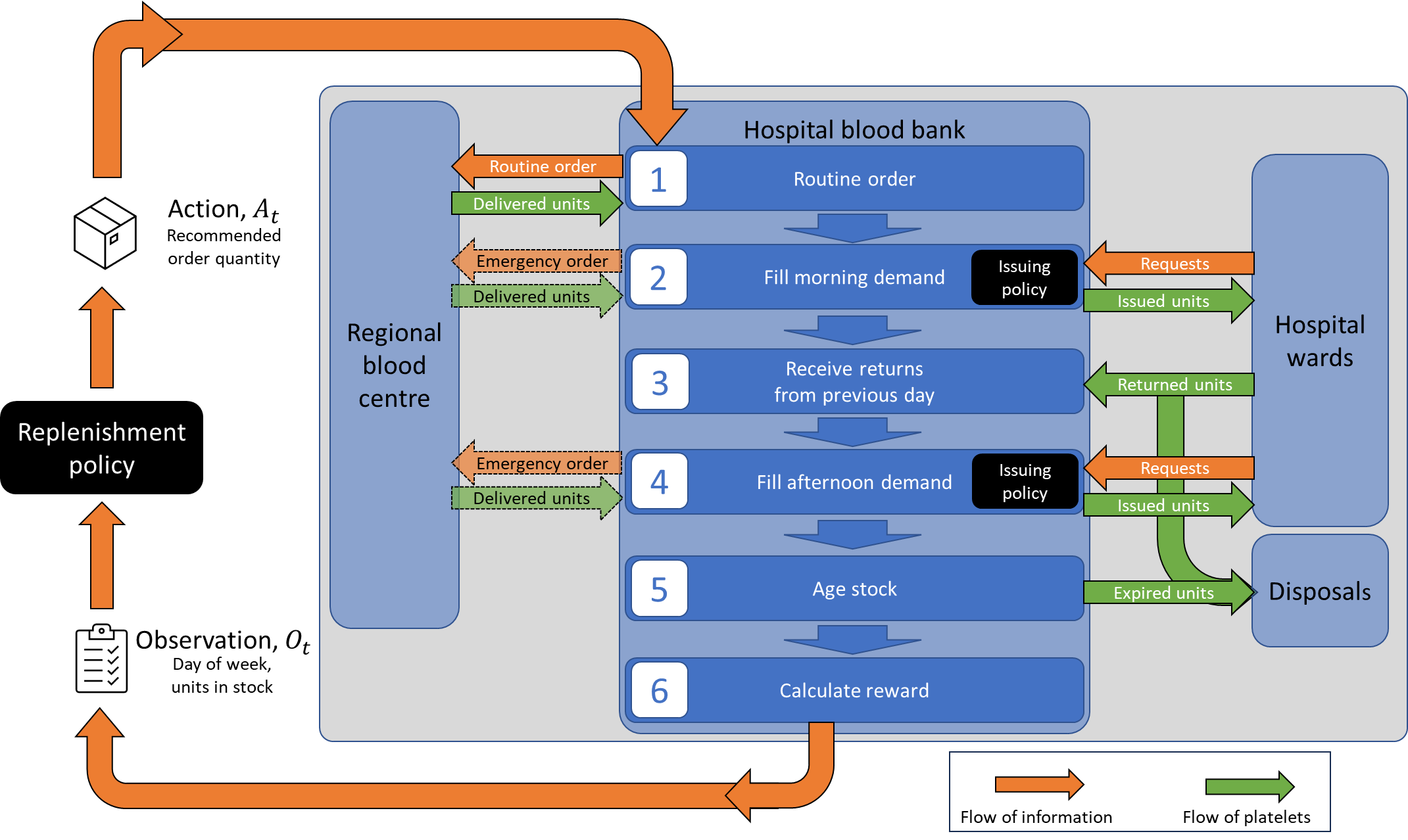}
\caption{\textbf{The order of events in one step of the simulated workflow, corresponding to one day.} Emergency orders are only placed at stages 2 and 4 if there is insufficient stock to meet demand, and are made for one request at a time.}\label{fig:sim_steps}
\end{figure}

The simulation we developed of the workflow for managing platelets in a hospital blood bank is illustrated in Figure \ref{fig:sim_steps}, showing the six stages that occur each day. A routine order is placed each morning based on the replenishment policy and is assumed to arrive immediately from the regional blood centre. Platelets have a fixed, known useful life $m$ days but do not always arrive fresh.  Demand is stochastic and requests for units from hospital wards are filled from available stock following the issuing policy during the morning and afternoon. Emergency orders are placed in the event of a shortage, which incurs a penalty. Units issued but not transfused on the previous day are returned at midday. Returned units that have expired since issue and units subject to slippage are discarded while other returned units are available to fill afternoon demand. We include slippage in the system to model potential problems with handling or storage such that not all units returned before expiry can be reissued \cite{kron_multicenter_2021}. Stock is aged at the end of day and expired units held in stock are discarded.   

\subsubsection{Overview of the simulated workflow}\label{sec:results:workflow}

The replenishment problem is framed as a Markov decision process (MDP), and implemented as a reinforcement learning environment using the Python library gymnax \cite{lange_gymnax_2022}. An MDP models a problem in which an agent makes a decision at a discrete series of points in time $t$. At each timestep (a day in our simulation), the agent observes the state of its environment, $S_t$, and takes an action $A_t$. The environment may change in response, and at the next point in time, $t+1$, the agent will receive a reward signal $R_{t+1}$ and the updated state of the environment $S_{t+1}$, and must select its next action $A_t$. The goal is to learn a policy, a function that maps from observations of the state to an action, that maximises the expected sum of discounted future rewards. In a problem with an infinite horizon the expected sum of discounted future rewards at timestep $t$, also called the return, is $G_t = \sum^{\infty}_{k=0}\gamma^kR_{t+k+1}$, where the discount factor $\gamma$ defines the relative importance of immediate and delayed rewards. 

The reward function comprises five components: a holding cost per unit in stock at the end of the period ($C_h=130$), a shortage cost per unit of unmet demand ($C_s=3,250$), a wastage cost per unit that perishes or is lost to slippage ($C_w=650$), a variable ordering cost per unit purchased ($C_v=650$) and a fixed ordering cost which is incurred when $A_t > 0$ ($C_f=225$). The elements of the reward function are based on those those used by Rajendran and Ravindran \cite{rajendran_platelet_2017}, gathered from a regional medical centre in Pennsylvania, USA. The ratio of 1:5 between the cost of a wasted unit and a shortage is commonly seen in platelet inventory management problems \cite{haijema_blood_2007, van_dijk_blood_2009}. We use a discount factor $\gamma=1$ in all of our experiments and therefore report the mean daily cost (the negative mean daily reward) instead of the return $G$ for an episode. With $\gamma=1$ these measures of performance are directly proportional, and we consider the mean daily cost easier to interpret. 

Demand follows a Poisson distribution and we assume that half of the demand arises in the morning (at stage 2) and half the the demand arises in the afternoon (at stage 4). The four key inputs to the simulation are the mean daily demand $\mu_{\tau}, \forall \tau \in \{0, 1, ..., 6\}$; the return rate $\rho$; the slippage rate $\phi$; and the parameters for the multinomial distribution of the remaining useful life on arrival for each weekday $\underline{\Delta}_{\tau}, \forall \tau \in \{0, 1, ..., 6\}$. We estimated values for each of these inputs using data from the UCLH transfusion laboratory information system Bank Manager for the years 2015 and 2016. The maximum useful life on arrival, $m$, is 5 days, in line with the majority of units received by the UCLH transfusion laboratory during this period. See Supplementary Note \ref{app:sim_processes} for the additional details on the simulated workflow and Supplementary Note \ref{app:sim_inputs} for a detailed description of how the inputs were estimated from the underlying data. We also considered an alternative distribution of remaining useful life on arrival, with $m=3$, previously reported by Rajendran and Ravindran \cite{rajendran_platelet_2017}. 

\subsubsection{Replenishment policy}\label{sec:rep_policy}

The action taken in the MDP is the daily order quantity. We do not focus on replenishment policies in this work and therefore consider two straightforward, heuristic replenishment policies: a standing order policy and an $(\texttt{s}, \texttt{S})$ policy. The standing order policy is similar to the existing replenishment policy at UCLH (excluding additional orders later in the day): a fixed number of units are ordered each day. In an $(\texttt{s}, \texttt{S})$, an order is placed to bring stock up to the order-up-to level, $\texttt{S}$, if the current stock level is less than or equal to the re-order point $\texttt{s}$. We found an $(\texttt{s}, \texttt{S})$ policy to be near optimal in previous work considering a platelet replenishment scenario described by Mirjalili \cite{mirjalili_data-driven_2022}, in which demand also depended on the day of the week and not all units arrived fresh \cite{farrington_going_2023}. These policies are unlikely to be optimal because they do not take into account the age profile of the stock \cite{blake_use_2009}, or any information about the units issued on the previous day that will be returned, but they provide good baselines, for which parameters can be fit in a reasonable amount of time, that enabled us to explore our main interest: the potential of an ML-guided issuing policy. 

Following the approach described in our previous work \cite{farrington_going_2023}, we fit the parameters for the replenishment policies using simulation optimization, using the Python library Optuna \cite{akiba_optuna_2019} to suggest candidate sets of parameters. See Supplementary Material \ref{app:rep_policies} for additional information about the replenishment policies and how we fit the parameters. 

\subsubsection{Issuing policy} \label{sec:issuing_policy}

Our proposed issuing policy assumes that we have access to an ML model that makes a binary prediction for each request: will the requested platelets be returned to the hospital blood bank, or will they be transfused? If the model predicts that at least one unit in the request will be returned, we issue the freshest unit(s). Otherwise, we issue the oldest unit(s). In the simulation experiments, we make the assumption that the total demand comprises only requests for a single unit.  

The effectiveness of this approach depends on the quality of the predictive model. Two metrics for assessing the performance of a binary classification model are the sensitivity (the proportion of requests where the platelets were returned that the model predicted would be returned) and the specificity (the proportion of requests where the platelets were transfused that the model predicted would be transfused). If the model had a sensitivity = 0.0 and a specificity = 1.0, it would predict that all requested units would be transfused and our policy would be equivalent to an unmodified OUFO policy. Alternatively, if the model had a sensitivity = 1.0 and a specificity = 0.0, it would predict that all requested units would be returned and our policy would be equivalent to an unmodified Youngest-Unit First-Out policy.

In a conventional ML workflow, these metrics would be calculated on a portion of the dataset held-out from training, with known outcomes, to estimate the generalization performance of the model. Here, for our simulation-first approach, we need to simulate both the true labels (i.e. will the requested unit be returned or transfused) and the predicted label. The true label for each request is sampled from a binomial distribution with a probability of success equal to the return rate $\rho$. To generate the predicted label, we specified levels of sensitivity ($\alpha$) and specificity ($\beta$) and simulated the predictions that would be made by a model with that level of performance. If the sampled true label was 1 (and therefore the unit would not be transfused),  the predicted label was set to 1 if a sample drawn from a uniform distribution between 0 and 1 was less than $\alpha$ and 0 otherwise. If the sampled true label was 0, the predicted label was set to 1 if a sample from a uniform distribution between 0 and 1 was greater than $\beta$, and 0 otherwise. This process is described in psuedocode in Algorithm 1 in Supplementary Note \ref{app:sim_processes}. There is a temporal order to requests within a period, and the issuing policy was applied to one request at a time in this order. 

\subsubsection{Initial simulation experiments}\label{methods:sim:sim_exp}

We summarise the settings for the initial simulation experiments and sensitivity analyses, Experiments 1--7, in Table \ref{tab:sim_exp_summary} in Supplementary Note \ref{app:sim_inputs}. Experiments 1 and 2 represent a setting with returns, using a a standing order and $(\texttt{s}, \texttt{S})$ replenishment policy respectively. The mean daily demand was estimated from UCLH data including returns (UCLH Tx+r), as set out in Table \ref{tab:demand_inc_returns} in Supplementary Note \ref{app:sim_inputs}. We used $\rho = 8\%$ and $\phi=7\%$, estimated based on data from UCLH as described in Supplementary Note \ref{app:sim_inputs} and the parameters for the distribution of remaining useful life on arrival estimated from UCLH data set out in Table \ref{tab:ul_on_arrival_uclh} in Supplementary Note \ref{app:sim_inputs}. d the potential benefit of our proposed ML based issuing policy, YUPR, to understand whether it could be beneficial and, if so, how good the model would need to be to improve KPIs. In each experiment, we considered all pairs of predictive model sensitivity and specificity between 0 and 1 in increments of 0.1. This range includes a policy that is equivalent to an OUFO policy (when $\text{sensitivity} = 0.0$ and $\text{specificity} = 1.0$) and a perfect predictive model (PPM; when $\text{sensitivity} = 1.0$ and $\text{specificity} = 1.0$). Comparing the results for the YUPR policy with a simulated PPM to a baseline OUFO policy tells us whether the proposed policy could outperform OUFO, while comparing intermediate values to the OUFO policy identifies a region of combinations of predictive model sensitivity and specificity where the proposed issuing policy could provide a benefit. 

The benefit of our proposed issuing policy may depend on inputs to the simulated workflow. We have not identified any previously published estimates of the return rate or slippage rate, but Rajendran and Ravindran \cite{rajendran_platelet_2017} reported an alternative distribution of remaining useful life on arrival from a regional medical centre in Pennsylvania, USA (R\&R). The maximum useful life is three days, 50\% of the stock arrives fresh, 20\% with two days of useful life, and 30\% on the day that it will expire. The reason for this difference is that bacterial screening procedures adopted in the UK and elsewhere in Europe enable a shelf life of up to 7 days after donation, while until recently the US Food and Drug Administration limited storage to 5 days after donation  \cite{prioli_economic_2022, szczepiorkowski_platelet_2022}. We repeated Experiments 1 and 2, using the same settings except for this alternative distribution of remaining useful life on arrival, and report the results as Experiments 3 and 4. 

In each of Experiments 1-4, and for each issuing policy in experiments where the sensitivity and specificity of the predictive model were changed, we evaluated the combination of issuing policy and replenishment policy with the best parameters identified from simulation optimization on 10,000 rollouts, each 365 days long with a warm-up period of 100 days. We report the mean daily cost, mean wastage and mean service level over these 10,000 rollouts. For each rollout, the wastage was calculated as proportion of units received (from both routine and emergency orders) that were wasted (due to either expiry or slippage) over the 365 days following the warm-up period. The service level was calculated as the proportion of the total demand filled by units in stock (rather than requiring an emergency order) over the 365 days following the warm-up period. We also calculated the standard deviation of these metrics over the 10,000 evaluation rollouts which reflects the stochasticity in each scenario (due to random demand, remaining useful life, slippage, and whether a requested unit is transfused or returned).

When the only difference between two experiments is the policies, it is possible to use a paired-sample approach to directly compare their performance on each evaluation rollout because we used random seeds to generate the random elements of the simulation in a reproducible manner. For pairs of replenishment and issuing policy combinations in Experiments 1 and 2 (and, separately, Experiments 3 and 4), we calculated the difference in daily cost, service level and wastage achieved by the policies on each evaluation rollout, and then computed the mean of the differences and the standard error of the mean of the differences. These measures provide as estimate of the possible improvement in a metric by changing the replenishment policy alone, the issuing policy alone, or both policies. 

The results for Experiments 1--4 are presented in Tables \ref{tab:basic_sim} and \ref{tab:basic_sim_pairwise}, and in Figure \ref{fig:sim_contour}, in Supplementary Note \ref{app:sim_exp_results_all}. The results from Experiments 2 and 4 were used to generate the contour plots presented in Figure \ref{fig:contour_test}, to set the classification threshold and to evaluate the trained predictive model in terms of key performance indicators assuming the distribution of remaining useful life observed at UCLH and in a US hospital by Rajendran and Ravindran \cite{rajendran_platelet_2017}, respectively.

\subsubsection{Sensitivity analyses}

To evaluate the impact of the return rate and the slippage rate, and a wider range of distributions of remaining useful life on arrival, we conducted Experiments 5, 6 and 7 to vary each of these input parameters in turn. The other settings remained the same as in Experiment 2. We recorded the mean daily reward, wastage and service level when using an OUFO issuing policy and when using our proposed YUPR issuing policy with a PPM.  

In Experiment 5, we changed the return rate $\rho$ between 0.0 and 0.5 in increments of 0.05. We adjusted the mean daily demand as we adjusted the return rate so that the expected number of transfused units remained the same. The mean daily demand for each experiment is therefore equal to the mean daily demand for transfused units only reported in Table \ref{tab:demand_exc_returns} of Supplementary Note \ref{app:sim_inputs} (UCLH Tx) divided by by $(1-\rho)$. 

In Experiment 6, we changed the slippage rate $\phi$ between 0.0 and 1.0 in increments of 0.1. 

In Experiment 7, we changed the parameters of the multinomial distribution of remaining useful life on arrival. We use the same distribution for each day of week, with the parameters set equal to the values of a binomial distribution with 4 trials and a probability of success between 0.0 and 1.0 in increments of 0.1. When the probability of success is 1.0, all units arrive with five days of remaining useful life, and when the probability of success is 0.0 all units arrive on the day that they expire. The distributions for each value of the probability of success are set out in Table \ref{tab:ul_on_arrival_scen} in Supplementary Note \ref{app:sim_inputs}. 

Experiments 5--7 followed the same evaluation process as Experiments 1--4, and the paired-sample approach was used to compare the policies at each level of each changed input setting. 

The results for Experiments 5--7 are presented in Figures \ref{fig:scenario_plot} and \ref{fig:pairwise_scenario_plot} in Supplementary Note \ref{app:sim_exp_results_all}. 

\subsection{Hardware}
Data processing, model training and evaluation were conducted on a Windows-based secure virtual research environment. 

All simulation experiments were conducted on a desktop computer running Ubuntu 20.04 LTS via Windows Subsystem for Linux on Windows 11 with an AMD Ryzen 9 5900 X processor, 64GB RAM, and an Nvidia GeForce RTX 3060 GPU. 

\subsection{Ethical approval}
This work is part of the `TransfuseAI' project, which was sponsored by the UCLH/UCL Joint Research Office and received ethical approval from the NHS Health Research Authority on 1 March 2021 (IRAS project ID 290615).

\section{Data Availability} \label{sec:data_availability}

The datasets required to replicate the simulation experiments are included in the published article and its supplementary information files. The underlying data used to estimate the simulation inputs, and to train and evaluate the ML model, are not publicly available. Researchers wishing to validate or replicate this work using the same datasets would need to be approved for research collaborations with University College London Hospitals NHS Foundation Trust, and to secure appropriate permissions from the UCLH/UCL Joint Research Office. Researchers who meet these requirements can contact the corresponding author for further information about access to the datasets.

\section{Code Availability} \label{sec:code_availability}

The underlying code for this study is available on GitHub and can be accessed at \url{https://github.com/joefarrington/plt\_returns}. 

\backmatter

\bmhead{Supplementary information}

Additional information is provided in Supplementary Notes A--K.

\bmhead{Acknowledgements}

JF is funded by UKRI training grant EP/S021612/1, the CDT in AI-enabled Healthcare Systems. This study
was funded by the Clinical and Research Informatics Unit at the National Institute for Health and Care
Research University College London Hospitals Biomedical Research Centre. The funders played no role in study design, data collection, analysis and interpretation of data, or the writing of this manuscript. For the purpose of open access, the author has applied a Creative Commons Attribution (CC BY) licence to any Author Accepted Manuscript version arising.

The authors gratefully acknowledge the support of Roma Klapaukh, Saidul Haque, Dave Ramlakhan and Richard Clarke in identifying and extracting the data, and of the UCLH transfusion laboratory, particularly Ian Longair, in understanding the local systems, processes and challenges. The authors also thank Zella King and Andre Vauvelle for their insightful comments during the course of this project. 

\bmhead{Author contributions} \textbf{JF}: Conceptualization, Methodology, Software,  Investigation, Writing - Original Draft, Visualization. \textbf{SA}: Writing - Review \& Editing. \textbf{MU}: Conceptualization, Methodology, Writing - Review \& Editing, Visualization, Supervision. \textbf{KL}: Conceptualization, Methodology, Writing - Review \& Editing, Supervision, Project administration, Funding acquisition. \textbf{WKW}: Conceptualization, Resources, Supervision, Funding acquisition. All authors read and approved the  
final manuscript.
\bmhead{Competing interests}

All authors declare no financial or non-financial competing interests. 

%\section*{Declarations}

%Some journals require declarations to be submitted in a standardised format. Please check the Instructions for Authors of the journal to which you are submitting to see if you need to complete this section. If yes, your manuscript must contain the following sections under the heading `Declarations':

%\begin{itemize}
%\item Conflict of interest/Competing interests (check journal-specific guidelines for which heading to use)
%\item Ethics approval 
%\item Consent for publication
%\item Availability of data and materials
%\item Authors' contributions

%\noindent
%If any of the sections are not relevant to your manuscript, please include the heading and write `Not applicable' for that section. 

%\bigskip

%\bigskip\noindent
%Springer journals and proceedings: \url{https://www.springer.com/gp/editorial-policies}

%\bigskip\noindent
%Nature Portfolio journals: \url{https://www.nature.com/nature-research/editorial-policies}

%\bigskip\noindent
%\textit{Scientific Reports}: \url{https://www.nature.com/srep/journal-policies/editorial-policies}

%\bigskip\noindent
%BMC journals: \url{https://www.biomedcentral.com/getpublished/editorial-policies}
%\end{flushleft}

\pagebreak

\bibliography{returns_bib.bib}% common bib file

\begin{thebibliography}{100}
\expandafter\ifx\csname url\endcsname\relax
  \def\url#1{\burl{#1}}\fi
\expandafter\ifx\csname urlprefix\endcsname\relax\def\urlprefix{}\fi
\providecommand{\bibinfo}[2]{#2}
\providecommand{\eprint}[2][]{\url{#2}}
\providecommand{\doi}[1]{\url{https://doi.org/#1}}
\bibcommenthead

\bibitem{flint_is_2020}
\bibinfo{author}{Flint, A.~W.} \emph{et~al.}
\newblock \bibinfo{title}{Is platelet expiring out of date? {A} systematic review}.
\newblock \emph{\bibinfo{journal}{Transfus. Med. Rev.}} \textbf{\bibinfo{volume}{34}}, \bibinfo{pages}{42--50} (\bibinfo{year}{2020}).

\bibitem{blood_stocks_management_scheme_nhsbt_2018}
\bibinfo{author}{{Blood Stocks Management Scheme}}.
\newblock \bibinfo{title}{{NHSBT} hospital blood supply chain annual report 2017/18}.
\newblock \bibinfo{type}{Tech. Rep.} (\bibinfo{year}{2018}).
\newblock \urlprefix\url{https://nhsbtdbe.blob.core.windows.net/umbraco-assets-corp/15951/nhsbt-annual-report-2017-18.pdf}.

\bibitem{blood_stocks_management_scheme_2023}
\bibinfo{author}{{Blood Stocks Management Scheme}}.
\newblock \bibinfo{title}{2023 {BSMS} 10 year component review}.
\newblock \bibinfo{type}{Tech. Rep.} (\bibinfo{year}{2023}).
\newblock \urlprefix\url{https://nhsbtdbe.blob.core.windows.net/umbraco-assets-corp/31272/bsms-10-year-component-review-2.pdf}.

\bibitem{Yates2017}
\bibinfo{author}{Yates, N.}, \bibinfo{author}{Stanger, S.}, \bibinfo{author}{Wilding, R.} \& \bibinfo{author}{Cotton, S.}
\newblock \bibinfo{title}{Approaches to assessing and minimizing blood wastage in the hospital and blood supply chain}.
\newblock \emph{\bibinfo{journal}{ISBT Sci. Ser.}} \textbf{\bibinfo{volume}{12}}, \bibinfo{pages}{91--98} (\bibinfo{year}{2017}).

\bibitem{Stanger2012}
\bibinfo{author}{Stanger, S.~H.}, \bibinfo{author}{Yates, N.}, \bibinfo{author}{Wilding, R.} \& \bibinfo{author}{Cotton, S.}
\newblock \bibinfo{title}{Blood inventory management: hospital best practice}.
\newblock \emph{\bibinfo{journal}{Transfus. Med. Rev.}} \textbf{\bibinfo{volume}{26}}, \bibinfo{pages}{153--163} (\bibinfo{year}{2012}).

\bibitem{tissot_ethics_2016}
\bibinfo{author}{Tissot, J.-D.} \& \bibinfo{author}{Garraud, O.}
\newblock \bibinfo{title}{Ethics and blood donation: A marriage of convenience}.
\newblock \emph{\bibinfo{journal}{Presse Med.}} \textbf{\bibinfo{volume}{45}}, \bibinfo{pages}{e247--e252} (\bibinfo{year}{2016}).

\bibitem{Dumkreiger2020}
\bibinfo{author}{Dumkreiger, G.}
\newblock \emph{\bibinfo{title}{Data driven personalized management of hospital inventory of perishable and substitutable blood products}}.
\newblock Ph.D. thesis, \bibinfo{school}{Arizona State University}, \bibinfo{address}{Phoenix, AZ, USA} (\bibinfo{year}{2020}).
\newblock \urlprefix\url{https://keep.lib.asu.edu/items/158661}.

\bibitem{Osorio2015}
\bibinfo{author}{Osorio, A.~F.}, \bibinfo{author}{Brailsford, S.~C.} \& \bibinfo{author}{Smith, H.~K.}
\newblock \bibinfo{title}{A structured review of quantitative models in the blood supply chain: a taxonomic framework for decision-making}.
\newblock \emph{\bibinfo{journal}{Int. J. Prod. Econ.}} \textbf{\bibinfo{volume}{53}}, \bibinfo{pages}{7191--7212} (\bibinfo{year}{2015}).

\bibitem{Piraban2019}
\bibinfo{author}{Pirabán, A.}, \bibinfo{author}{Guerrero, W.~J.} \& \bibinfo{author}{Labadie, N.}
\newblock \bibinfo{title}{Survey on blood supply chain management: models and methods}.
\newblock \emph{\bibinfo{journal}{Comput. Oper. Res.}} \textbf{\bibinfo{volume}{112}} (\bibinfo{year}{2019}).

\bibitem{torrado_towards_2022}
\bibinfo{author}{Torrado, A.} \& \bibinfo{author}{Barbosa-Póvoa, A.}
\newblock \bibinfo{title}{Towards an {Optimized} and {Sustainable} {Blood} {Supply} {Chain} {Network} under {Uncertainty}: {A} {Literature} {Review}}.
\newblock \emph{\bibinfo{journal}{Clean. Logist. Supply Chain}} \textbf{\bibinfo{volume}{3}}, \bibinfo{pages}{100028} (\bibinfo{year}{2022}).

\bibitem{haijema_optimal_2014}
\bibinfo{author}{Haijema, R.}
\newblock \bibinfo{title}{Optimal ordering, issuance and disposal policies for inventory management of perishable products}.
\newblock \emph{\bibinfo{journal}{Int. J. Prod. Econ.}} \textbf{\bibinfo{volume}{157}}, \bibinfo{pages}{158--169} (\bibinfo{year}{2014}).

\bibitem{stanger_what_2012}
\bibinfo{author}{Stanger, S.~H.}, \bibinfo{author}{Wilding, R.}, \bibinfo{author}{Yates, N.} \& \bibinfo{author}{Cotton, S.}
\newblock \bibinfo{title}{What drives perishable inventory management performance? {Lessons} learnt from the {UK} blood supply chain}.
\newblock \emph{\bibinfo{journal}{Supply Chain Manag.}} \textbf{\bibinfo{volume}{17}}, \bibinfo{pages}{107--123} (\bibinfo{year}{2012}).

\bibitem{bakker_review_2012}
\bibinfo{author}{Bakker, M.}, \bibinfo{author}{Riezebos, J.} \& \bibinfo{author}{Teunter, R.~H.}
\newblock \bibinfo{title}{Review of inventory systems with deterioration since 2001}.
\newblock \emph{\bibinfo{journal}{Eur. J. Oper. Res.}} \textbf{\bibinfo{volume}{221}}, \bibinfo{pages}{275--284} (\bibinfo{year}{2012}).

\bibitem{chaudhary_state---art_2018}
\bibinfo{author}{Chaudhary, V.}, \bibinfo{author}{Kulshrestha, R.} \& \bibinfo{author}{Routroy, S.}
\newblock \bibinfo{title}{State-of-the-art literature review on inventory models for perishable products}.
\newblock \emph{\bibinfo{journal}{J. Adv. Manag. Res.}} \textbf{\bibinfo{volume}{15}}, \bibinfo{pages}{306--346} (\bibinfo{year}{2018}).

\bibitem{janssen_literature_2016}
\bibinfo{author}{Janssen, L.}, \bibinfo{author}{Claus, T.} \& \bibinfo{author}{Sauer, J.}
\newblock \bibinfo{title}{Literature review of deteriorating inventory models by key topics from 2012 to 2015}.
\newblock \emph{\bibinfo{journal}{Int. J. Prod. Econ.}} \textbf{\bibinfo{volume}{182}}, \bibinfo{pages}{86--112} (\bibinfo{year}{2016}).

\bibitem{karaesmen_managing_2011}
\bibinfo{author}{Karaesmen, I.~Z.}, \bibinfo{author}{Scheller–Wolf, A.} \& \bibinfo{author}{Deniz, B.}
\newblock \bibinfo{title}{ in \textit{Managing {Perishable} and {Aging} {Inventories}: {Review} and {Future} {Research} {Directions}}} (eds \bibinfo{editor}{Kempf, K.~G.}, \bibinfo{editor}{Keskinocak, P.} \& \bibinfo{editor}{Uzsoy, R.}) \emph{\bibinfo{booktitle}{Planning {Production} and {Inventories} in the {Extended} {Enterprise}: {A} {State} of the {Art} {Handbook}, {Volume} 1}} \bibinfo{pages}{393--436} (\bibinfo{publisher}{Springer US}, \bibinfo{year}{2011}).

\bibitem{mirabelli_optimization_2022}
\bibinfo{author}{Mirabelli, G.} \& \bibinfo{author}{Solina, V.}
\newblock \bibinfo{title}{Optimization strategies for the integrated management of perishable supply chains: {A} literature review}.
\newblock \emph{\bibinfo{journal}{J. Ind. Eng. Manag.}} \textbf{\bibinfo{volume}{15}}, \bibinfo{pages}{58--91} (\bibinfo{year}{2022}).

\bibitem{Nahmias1982}
\bibinfo{author}{Nahmias, S.}
\newblock \bibinfo{title}{Perishable inventory theory: a review}.
\newblock \emph{\bibinfo{journal}{Oper. Res.}} \textbf{\bibinfo{volume}{30}}, \bibinfo{pages}{680--708} (\bibinfo{year}{1982}).

\bibitem{pierskalla_optimal_1972}
\bibinfo{author}{Pierskalla, W.~P.} \& \bibinfo{author}{Roach, C.~D.}
\newblock \bibinfo{title}{Optimal {Issuing} {Policies} for {Perishable} {Inventory}}.
\newblock \emph{\bibinfo{journal}{Manage. Sci.}} \textbf{\bibinfo{volume}{18}}, \bibinfo{pages}{603--614} (\bibinfo{year}{1972}).

\bibitem{dalalah_platelets_2019}
\bibinfo{author}{Dalalah, D.}, \bibinfo{author}{Bataineh, O.} \& \bibinfo{author}{Alkhaledi, K.~A.}
\newblock \bibinfo{title}{Platelets inventory management: a rolling horizon sim–opt approach for an age-differentiated demand}.
\newblock \emph{\bibinfo{journal}{J. Simul.}} \textbf{\bibinfo{volume}{13}}, \bibinfo{pages}{209--225} (\bibinfo{year}{2019}).

\bibitem{civelek_blood_2015}
\bibinfo{author}{Civelek, I.}, \bibinfo{author}{Karaesmen, I.} \& \bibinfo{author}{Scheller-Wolf, A.}
\newblock \bibinfo{title}{Blood platelet inventory management with protection levels}.
\newblock \emph{\bibinfo{journal}{Eur. J. Oper. Res.}} \textbf{\bibinfo{volume}{243}}, \bibinfo{pages}{826--838} (\bibinfo{year}{2015}).

\bibitem{Dillon2017}
\bibinfo{author}{Dillon, M.}, \bibinfo{author}{Oliveira, F.} \& \bibinfo{author}{Abbasi, B.}
\newblock \bibinfo{title}{A two-stage stochastic programming model for inventory management in the blood supply chain}.
\newblock \emph{\bibinfo{journal}{Int. J. Prod. Econ.}} \textbf{\bibinfo{volume}{187}}, \bibinfo{pages}{27--41} (\bibinfo{year}{2017}).

\bibitem{duan_optimization_2014}
\bibinfo{author}{Duan, Q.} \& \bibinfo{author}{Liao, T.~W.}
\newblock \bibinfo{title}{Optimization of blood supply chain with shortened shelf lives and {ABO} compatibility}.
\newblock \emph{\bibinfo{journal}{Int. J. Prod. Econ.}} \textbf{\bibinfo{volume}{153}}, \bibinfo{pages}{113--129} (\bibinfo{year}{2014}).

\bibitem{meneses_blood_2021}
\bibinfo{author}{Meneses, M.}, \bibinfo{author}{Marques, I.} \& \bibinfo{author}{Barbosa-Póvoa, A.}
\newblock \bibinfo{title}{Blood inventory management: ordering policies for hospital blood banks under uncertainty}.
\newblock \emph{\bibinfo{journal}{Int. T. Oper. Res.}}  (\bibinfo{year}{2021}).

\bibitem{abbasi_issuing_2014}
\bibinfo{author}{Abbasi, B.} \& \bibinfo{author}{Hosseinifard, S.~Z.}
\newblock \bibinfo{title}{On the {Issuing} {Policies} for {Perishable} {Items} such as {Red} {Blood} {Cells} and {Platelets} in {Blood} {Service}}.
\newblock \emph{\bibinfo{journal}{Decision Sci.}} \textbf{\bibinfo{volume}{45}}, \bibinfo{pages}{995--1020} (\bibinfo{year}{2014}).

\bibitem{atkinson_novel_2012}
\bibinfo{author}{Atkinson, M.~P.}, \bibinfo{author}{Fontaine, M.~J.}, \bibinfo{author}{Goodnough, L.~T.} \& \bibinfo{author}{Wein, L.~M.}
\newblock \bibinfo{title}{A novel allocation strategy for blood transfusions: investigating the tradeoff between the age and availability of transfused blood}.
\newblock \emph{\bibinfo{journal}{Transfusion}} \textbf{\bibinfo{volume}{52}}, \bibinfo{pages}{108--117} (\bibinfo{year}{2012}).

\bibitem{Abdulwahab2014}
\bibinfo{author}{Abdulwahab, U.} \& \bibinfo{author}{Wahab, M.~I.}
\newblock \bibinfo{title}{Approximate dynamic programming modeling for a typical blood platelet bank}.
\newblock \emph{\bibinfo{journal}{Comput. Ind. Eng.}} \textbf{\bibinfo{volume}{78}}, \bibinfo{pages}{259--270} (\bibinfo{year}{2014}).

\bibitem{staves_electronic_2008}
\bibinfo{author}{Staves, J.} \emph{et~al.}
\newblock \bibinfo{title}{Electronic remote blood issue: a combination of remote blood issue with a system for end-to-end electronic control of transfusion to provide a “total solution” for a safe and timely hospital blood transfusion service}.
\newblock \emph{\bibinfo{journal}{Transfusion}} \textbf{\bibinfo{volume}{48}}, \bibinfo{pages}{415--424} (\bibinfo{year}{2008}).

\bibitem{jennings_analysis_1968}
\bibinfo{author}{Jennings, J.~B.}
\newblock \bibinfo{title}{An {Analysis} of {Hospital} {Blood} {Bank} {Whole} {Blood} {Inventory} {Control} {Policies}}.
\newblock \emph{\bibinfo{journal}{Transfusion}} \textbf{\bibinfo{volume}{8}}, \bibinfo{pages}{335--342} (\bibinfo{year}{1968}).

\bibitem{pereira_blood_2005}
\bibinfo{author}{Pereira, A.}
\newblock \bibinfo{title}{Blood inventory management in the type and screen era}.
\newblock \emph{\bibinfo{journal}{Vox Sang.}} \textbf{\bibinfo{volume}{89}}, \bibinfo{pages}{245--250} (\bibinfo{year}{2005}).

\bibitem{rytila_using_2006}
\bibinfo{author}{Rytilä, J.~S.} \& \bibinfo{author}{Spens, K.~M.}
\newblock \bibinfo{title}{Using simulation to increase efficiency in blood supply chains}.
\newblock \emph{\bibinfo{journal}{Manag. Res. News}} \textbf{\bibinfo{volume}{29}}, \bibinfo{pages}{801--819} (\bibinfo{year}{2006}).

\bibitem{Katsaliaki2007}
\bibinfo{author}{Katsaliaki, K.} \& \bibinfo{author}{Brailsford, S.~C.}
\newblock \bibinfo{title}{Using simulation to improve the blood supply chain}.
\newblock \emph{\bibinfo{journal}{J. Oper. Res. Soc.}} \textbf{\bibinfo{volume}{58}}, \bibinfo{pages}{219--227} (\bibinfo{year}{2007}).

\bibitem{yahnke_analysis_1972}
\bibinfo{author}{Yahnke, D.~P.}, \bibinfo{author}{Rimm, A.~A.}, \bibinfo{author}{Mundt, C.~J.}, \bibinfo{author}{Aster, R.~H.} \& \bibinfo{author}{Hurst, T.~M.}
\newblock \bibinfo{title}{Analysis and {Optimization} of a {Regional} {Blood} {Bank} {Distribution} {Process}}.
\newblock \emph{\bibinfo{journal}{Transfusion}} \textbf{\bibinfo{volume}{12}}, \bibinfo{pages}{111--118} (\bibinfo{year}{1972}).

\bibitem{yahnke_analysis_1973}
\bibinfo{author}{Yahnke, D.~P.}, \bibinfo{author}{Rimm, A.~A.}, \bibinfo{author}{Makowski, G.~G.} \& \bibinfo{author}{Aster, R.~H.}
\newblock \bibinfo{title}{Analysis and {Optimization} of a {Regional} {Blood} {Bank} {Distribution} {Process}: {II}. {Derivation} and {Use} of a {Method} for {Evaluating} {Hospital} {Management} {Procedures}}.
\newblock \emph{\bibinfo{journal}{Transfusion}} \textbf{\bibinfo{volume}{13}}, \bibinfo{pages}{156--169} (\bibinfo{year}{1973}).

\bibitem{britten_weekly_1979}
\bibinfo{author}{Britten, A. F.~H.} \& \bibinfo{author}{Geurtze, D.~G.}
\newblock \bibinfo{title}{Weekly {Rotation} of {Blood} {Inventory}—{A} {System} for {Supplying} {Small} {Hospitals}}.
\newblock \emph{\bibinfo{journal}{Transfusion}} \textbf{\bibinfo{volume}{19}}, \bibinfo{pages}{738--741} (\bibinfo{year}{1979}).

\bibitem{chen_managing_2022}
\bibinfo{author}{Chen, K.}, \bibinfo{author}{Song, J.-S.}, \bibinfo{author}{Shang, J.} \& \bibinfo{author}{Xiao, T.}
\newblock \bibinfo{title}{Managing hospital platelet inventory with mid-cycle expedited replenishments and returns}.
\newblock \emph{\bibinfo{journal}{Prod. Oper. Manag.}} \textbf{\bibinfo{volume}{31}}, \bibinfo{pages}{2015--2037} (\bibinfo{year}{2022}).

\bibitem{ahmadimanesh_designing_2022}
\bibinfo{author}{Ahmadimanesh, M.}, \bibinfo{author}{Pooya, A.}, \bibinfo{author}{Safabakhsh, H.} \& \bibinfo{author}{Sadeghi, S.}
\newblock \bibinfo{title}{Designing an {Optimal} {Model} of {Blood} {Logistics} {Management} with the {Possibility} of {Return} in the {Three}-{Level} {Blood} {Transfusion} {Network}}.
\newblock \bibinfo{howpublished}{Research Square} (\bibinfo{year}{2022}).
\newblock \urlprefix\url{https://www.researchsquare.com/article/rs-1187827/v1}.

\bibitem{ambilkar_product_2021}
\bibinfo{author}{Ambilkar, P.}, \bibinfo{author}{Dohale, V.}, \bibinfo{author}{Gunasekaran, A.} \& \bibinfo{author}{Bilolikar, V.}
\newblock \bibinfo{title}{Product returns management: a comprehensive review and future research agenda}.
\newblock \emph{\bibinfo{journal}{Int. J. Prod. Econ.}} \textbf{\bibinfo{volume}{0}}, \bibinfo{pages}{1--25} (\bibinfo{year}{2021}).

\bibitem{abdulla_taking_2019}
\bibinfo{author}{Abdulla, H.}, \bibinfo{author}{Ketzenberg, M.} \& \bibinfo{author}{Abbey, J.~D.}
\newblock \bibinfo{title}{Taking stock of consumer returns: {A} review and classification of the literature}.
\newblock \emph{\bibinfo{journal}{J. Oper. Manag.}} \textbf{\bibinfo{volume}{65}}, \bibinfo{pages}{560--605} (\bibinfo{year}{2019}).

\bibitem{canda_modeling_2015}
\bibinfo{author}{Canda, A.}, \bibinfo{author}{Yuan, X.-M.} \& \bibinfo{author}{Wang, F.-Y.}
\newblock \bibinfo{title}{Modeling and forecasting product returns: {An} industry case study}.
\newblock \emph{\bibinfo{journal}{2015 {IEEE} {International} {Conference} on {Industrial} {Engineering} and {Engineering} {Management} ({IEEM})}} \bibinfo{pages}{871--875} (\bibinfo{year}{Singapore, 6-9 December 2015}).

\bibitem{tsiliyannis_markov_2018}
\bibinfo{author}{Tsiliyannis, C.~A.}
\newblock \bibinfo{title}{Markov chain modeling and forecasting of product returns in remanufacturing based on stock mean-age}.
\newblock \emph{\bibinfo{journal}{Eur. J. Oper. Res.}} \textbf{\bibinfo{volume}{271}}, \bibinfo{pages}{474--489} (\bibinfo{year}{2018}).

\bibitem{chou_policies_2020}
\bibinfo{author}{Chou, M.~C.}, \bibinfo{author}{Sim, C.-K.} \& \bibinfo{author}{Yuan, X.-M.}
\newblock \bibinfo{title}{Policies for inventory models with product returns forecast from past demands and past sales}.
\newblock \emph{\bibinfo{journal}{Ann. Oper. Res.}} \textbf{\bibinfo{volume}{288}}, \bibinfo{pages}{137--180} (\bibinfo{year}{2020}).

\bibitem{cui_predicting_2020}
\bibinfo{author}{Cui, H.}, \bibinfo{author}{Rajagopalan, S.} \& \bibinfo{author}{Ward, A.~R.}
\newblock \bibinfo{title}{Predicting product return volume using machine learning methods}.
\newblock \emph{\bibinfo{journal}{Eur. J. Oper. Res.}} \textbf{\bibinfo{volume}{281}}, \bibinfo{pages}{612--627} (\bibinfo{year}{2020}).

\bibitem{Joshi2019}
\bibinfo{author}{Joshi, I.} \& \bibinfo{author}{Morley, J.}
\newblock \bibinfo{title}{Artificial {Intelligence}: {How} to get it right}.
\newblock \bibinfo{howpublished}{NHS Transformation Directorate} (\bibinfo{year}{2019}).
\newblock \urlprefix\url{https://transform.england.nhs.uk/media/documents/NHSX_AI_report.pdf}.

\bibitem{zhu_local_2018}
\bibinfo{author}{Zhu, Y.}, \bibinfo{author}{Li, J.}, \bibinfo{author}{He, J.}, \bibinfo{author}{Quanz, B.~L.} \& \bibinfo{author}{Deshpande, A.~A.}
\newblock \bibinfo{title}{A {Local} {Algorithm} for {Product} {Return} {Prediction} in {E}-{Commerce}}.
\newblock \emph{\bibinfo{journal}{Proceedings of the {Twenty}-{Seventh} {International} {Joint} {Conference} on {Artificial} {Intelligence}}} \bibinfo{pages}{3718--3724} (\bibinfo{year}{Stockholm, Sweden, July 13-19 2018}).

\bibitem{li_trust-aware_2019}
\bibinfo{author}{Li, X.}, \bibinfo{author}{Zhuang, Y.}, \bibinfo{author}{Fu, Y.} \& \bibinfo{author}{He, X.}
\newblock \bibinfo{title}{A trust-aware random walk model for return propensity estimation and consumer anomaly scoring in online shopping}.
\newblock \emph{\bibinfo{journal}{Sci. China Inform. Sci.}} \textbf{\bibinfo{volume}{62}}, \bibinfo{pages}{52101} (\bibinfo{year}{2019}).

\bibitem{fu_fused_2016}
\bibinfo{author}{Fu, Y.} \emph{et~al.}
\newblock \bibinfo{title}{Fused latent models for assessing product return propensity in online commerce}.
\newblock \emph{\bibinfo{journal}{Decis. Support Syst.}} \textbf{\bibinfo{volume}{91}}, \bibinfo{pages}{77--88} (\bibinfo{year}{2016}).

\bibitem{brodheim_demand_1980}
\bibinfo{author}{Brodheim, E.} \& \bibinfo{author}{Prastacos, G.}
\newblock \bibinfo{title}{Demand, usage and issuing of blood at hospital blood banks}.
\newblock \bibinfo{type}{Tech. Rep.}, \bibinfo{institution}{Operations Research Laboratory, The New York Blood Center} (\bibinfo{year}{1980}).

\bibitem{shah_making_2019}
\bibinfo{author}{Shah, N.~H.}, \bibinfo{author}{Milstein, A.} \& \bibinfo{author}{Bagley, P., Steven~C.}
\newblock \bibinfo{title}{Making {Machine} {Learning} {Models} {Clinically} {Useful}}.
\newblock \emph{\bibinfo{journal}{JAMA, J. Am. Med. Assoc.}} \textbf{\bibinfo{volume}{322}}, \bibinfo{pages}{1351--1352} (\bibinfo{year}{2019}).

\bibitem{zhou_clinical_2021}
\bibinfo{author}{Zhou, Q.}, \bibinfo{author}{Chen, Z.-h.}, \bibinfo{author}{Cao, Y.-h.} \& \bibinfo{author}{Peng, S.}
\newblock \bibinfo{title}{Clinical impact and quality of randomized controlled trials involving interventions evaluating artificial intelligence prediction tools: a systematic review}.
\newblock \emph{\bibinfo{journal}{npj Digit. Med.}} \textbf{\bibinfo{volume}{4}}, \bibinfo{pages}{1--12} (\bibinfo{year}{2021}).

\bibitem{li_developing_2020}
\bibinfo{author}{Li, R.~C.}, \bibinfo{author}{Asch, S.~M.} \& \bibinfo{author}{Shah, N.~H.}
\newblock \bibinfo{title}{Developing a delivery science for artificial intelligence in healthcare}.
\newblock \emph{\bibinfo{journal}{npj Digit. Med.}} \textbf{\bibinfo{volume}{3}}, \bibinfo{pages}{1--3} (\bibinfo{year}{2020}).

\bibitem{misic_simulation-based_2021}
\bibinfo{author}{Mišić, V.~V.}, \bibinfo{author}{Rajaram, K.} \& \bibinfo{author}{Gabel, E.}
\newblock \bibinfo{title}{A simulation-based evaluation of machine learning models for clinical decision support: application and analysis using hospital readmission}.
\newblock \emph{\bibinfo{journal}{npj Digit. Med.}} \textbf{\bibinfo{volume}{4}}, \bibinfo{pages}{1--11} (\bibinfo{year}{2021}).

\bibitem{wornow_aplus_2023}
\bibinfo{author}{Wornow, M.}, \bibinfo{author}{Gyang~Ross, E.}, \bibinfo{author}{Callahan, A.} \& \bibinfo{author}{Shah, N.~H.}
\newblock \bibinfo{title}{{APLUS}: {A} {Python} library for usefulness simulations of machine learning models in healthcare}.
\newblock \emph{\bibinfo{journal}{J. Biomed. Inform.}} \textbf{\bibinfo{volume}{139}}, \bibinfo{pages}{104319} (\bibinfo{year}{2023}).

\bibitem{taylor_road_2021}
\bibinfo{author}{Taylor, J.~A.} \emph{et~al.}
\newblock \bibinfo{title}{The road to hell is paved with good intentions: the experience of applying for national data for linkage and suggestions for improvement}.
\newblock \emph{\bibinfo{journal}{BMJ Open}} \textbf{\bibinfo{volume}{11}}, \bibinfo{pages}{e047575} (\bibinfo{year}{2021}).

\bibitem{viet_value_2018}
\bibinfo{author}{Viet, N.~Q.}, \bibinfo{author}{Behdani, B.} \& \bibinfo{author}{Bloemhof, J.}
\newblock \bibinfo{title}{The value of information in supply chain decisions: {A} review of the literature and research agenda}.
\newblock \emph{\bibinfo{journal}{Comput. Ind. Eng.}} \textbf{\bibinfo{volume}{120}}, \bibinfo{pages}{68--82} (\bibinfo{year}{2018}).

\bibitem{Kopach2008}
\bibinfo{author}{Kopach, R.}, \bibinfo{author}{Balcioǧlu, B.} \& \bibinfo{author}{Carter, M.}
\newblock \bibinfo{title}{Tutorial on constructing a red blood cell inventory management system with two demand rates}.
\newblock \emph{\bibinfo{journal}{Eur. J. Oper. Res.}} \textbf{\bibinfo{volume}{185}}, \bibinfo{pages}{1051--1059} (\bibinfo{year}{2008}).

\bibitem{yuzgec_simulation_2013}
\bibinfo{author}{Yuzgec, E.}, \bibinfo{author}{Han, Y.} \& \bibinfo{author}{Nagarur, N.}
\newblock \bibinfo{title}{A simulation model for blood supply chain systems}.
\newblock \emph{\bibinfo{journal}{Proceedings of the 2013 {Industrial} and {Systems} {Engineering} {Research} {Conference}}} \bibinfo{pages}{1703--1711} (\bibinfo{year}{San Juan, Puerto Rico, 18-22 May 2013}).

\bibitem{katz_simulation_1983}
\bibinfo{author}{Katz, A.~J.}, \bibinfo{author}{Carter, C.~W.}, \bibinfo{author}{Saxton, P.}, \bibinfo{author}{Blutt, J.} \& \bibinfo{author}{Kakaiya, R.~M.}
\newblock \bibinfo{title}{Simulation analysis of platelet production and inventory management}.
\newblock \emph{\bibinfo{journal}{Vox Sang.}} \textbf{\bibinfo{volume}{44}}, \bibinfo{pages}{31--36} (\bibinfo{year}{1983}).

\bibitem{asllani_simulation-based_2014}
\bibinfo{author}{Asllani, A.}, \bibinfo{author}{Culler, E.} \& \bibinfo{author}{Ettkin, L.}
\newblock \bibinfo{title}{A simulation-based apheresis platelet inventory management model}.
\newblock \emph{\bibinfo{journal}{Transfusion}} \textbf{\bibinfo{volume}{54}}, \bibinfo{pages}{2730--2735} (\bibinfo{year}{2014}).

\bibitem{baesler_analysis_2014}
\bibinfo{author}{Baesler, F.}, \bibinfo{author}{Nemeth, M.}, \bibinfo{author}{Martínez, C.} \& \bibinfo{author}{Bastías, A.}
\newblock \bibinfo{title}{Analysis of inventory strategies for blood components in a regional blood center using process simulation}.
\newblock \emph{\bibinfo{journal}{Transfusion}} \textbf{\bibinfo{volume}{54}}, \bibinfo{pages}{323--330} (\bibinfo{year}{2014}).

\bibitem{rajendran_hybrid_2020}
\bibinfo{author}{Rajendran, S.} \& \bibinfo{author}{Srinivas, S.}
\newblock \bibinfo{title}{Hybrid ordering policies for platelet inventory management under demand uncertainty}.
\newblock \emph{\bibinfo{journal}{IISE Trans. Healthc. Syst. Eng.}} \textbf{\bibinfo{volume}{10}}, \bibinfo{pages}{113--126} (\bibinfo{year}{2020}).

\bibitem{duan_new_2013}
\bibinfo{author}{Duan, Q.} \& \bibinfo{author}{Liao, T.~W.}
\newblock \bibinfo{title}{A new age-based replenishment policy for supply chain inventory optimization of highly perishable products}.
\newblock \emph{\bibinfo{journal}{Int. J. Prod. Econ.}} \textbf{\bibinfo{volume}{145}}, \bibinfo{pages}{658--671} (\bibinfo{year}{2013}).

\bibitem{blake_optimizing_2003}
\bibinfo{author}{Blake, J.~T.} \emph{et~al.}
\newblock \bibinfo{title}{Optimizing the platelet supply chain in {Nova} {Scotia}}.
\newblock \emph{\bibinfo{journal}{Proceedings of the 29th meeting of the {European} {Working} {Group} on {Operational} {Research} {Applied} to {Health} {Services}}} \bibinfo{pages}{47--66} (\bibinfo{year}{Prague, Czech Republic, 27 July - 1 August 2003}).

\bibitem{haijema_blood_2007}
\bibinfo{author}{Haijema, R.}, \bibinfo{author}{van~der Wal, J.} \& \bibinfo{author}{van Dijk, N.~M.}
\newblock \bibinfo{title}{Blood platelet production: optimization by dynamic programming and simulation}.
\newblock \emph{\bibinfo{journal}{Comput. Oper. Res.}} \textbf{\bibinfo{volume}{34}}, \bibinfo{pages}{760--779} (\bibinfo{year}{2007}).

\bibitem{li_decision_2021}
\bibinfo{author}{Li, N.}, \bibinfo{author}{Chiang, F.}, \bibinfo{author}{Down, D.~G.} \& \bibinfo{author}{Heddle, N.~M.}
\newblock \bibinfo{title}{A decision integration strategy for short-term demand forecasting and ordering for red blood cell components}.
\newblock \emph{\bibinfo{journal}{Oper. Res. Health Care}} \textbf{\bibinfo{volume}{29}}, \bibinfo{pages}{100290} (\bibinfo{year}{2021}).

\bibitem{motamedi_demand_2021}
\bibinfo{author}{Motamedi, M.}, \bibinfo{author}{Li, N.}, \bibinfo{author}{Down, D.~G.} \& \bibinfo{author}{Heddle, N.~M.}
\newblock \bibinfo{title}{Demand forecasting for platelet usage: from univariate time series to multivariate models}.
\newblock \bibinfo{howpublished}{arXiv} (\bibinfo{year}{2021}).
\newblock \urlprefix\url{http://arxiv.org/abs/2101.02305}.

\bibitem{schilling_reduction_2022}
\bibinfo{author}{Schilling, M.}, \bibinfo{author}{Rickmann, L.}, \bibinfo{author}{Hutschenreuter, G.} \& \bibinfo{author}{Spreckelsen, C.}
\newblock \bibinfo{title}{Reduction of platelet outdating and shortage by forecasting demand with statistical learning and deep neural networks: modeling study}.
\newblock \emph{\bibinfo{journal}{JMIR Med. Inform.}} \textbf{\bibinfo{volume}{10}}, \bibinfo{pages}{e29978} (\bibinfo{year}{2022}).

\bibitem{Guan2017d}
\bibinfo{author}{Guan, L.} \emph{et~al.}
\newblock \bibinfo{title}{Big data modeling to predict platelet usage and minimize wastage in a tertiary care system}.
\newblock \emph{\bibinfo{journal}{P. Natl. Acad. Sci. USA}} \textbf{\bibinfo{volume}{114}}, \bibinfo{pages}{11368--11373} (\bibinfo{year}{2017}).

\bibitem{abouee_mehrizi_data-driven_2022}
\bibinfo{author}{Abouee-Mehrizi, H.}, \bibinfo{author}{Mirjalili, M.} \& \bibinfo{author}{Sarhangian, V.}
\newblock \bibinfo{title}{Data-driven platelet inventory management under uncertainty in the remaining shelf life of units}.
\newblock \emph{\bibinfo{journal}{Prod. Oper. Manag.}} \textbf{\bibinfo{volume}{31}}, \bibinfo{pages}{3914--3932} (\bibinfo{year}{2022}).

\bibitem{fildes_incorporating_2011}
\bibinfo{author}{Fildes, R.} \& \bibinfo{author}{Kingsman, B.}
\newblock \bibinfo{title}{Incorporating demand uncertainty and forecast error in supply chain planning models}.
\newblock \emph{\bibinfo{journal}{J. Oper. Res. Soc.}} \textbf{\bibinfo{volume}{62}}, \bibinfo{pages}{483--500} (\bibinfo{year}{2011}).

\bibitem{altendorfer_effects_2016}
\bibinfo{author}{Altendorfer, K.}, \bibinfo{author}{Felberbauer, T.} \& \bibinfo{author}{Jodlbauer, H.}
\newblock \bibinfo{title}{Effects of forecast errors on optimal utilisation in aggregate production planning with stochastic customer demand}.
\newblock \emph{\bibinfo{journal}{Int. J. Prod. Econ.}} \textbf{\bibinfo{volume}{54}}, \bibinfo{pages}{3718--3735} (\bibinfo{year}{2016}).

\bibitem{sanders_quantifying_2009}
\bibinfo{author}{Sanders, N.} \& \bibinfo{author}{Graman, G.}
\newblock \bibinfo{title}{Quantifying costs of forecast errors: {A} case study of the warehouse environment}.
\newblock \emph{\bibinfo{journal}{Omega}} \textbf{\bibinfo{volume}{37}}, \bibinfo{pages}{116--125} (\bibinfo{year}{2009}).

\bibitem{rajendran_platelet_2017}
\bibinfo{author}{Rajendran, S.} \& \bibinfo{author}{Ravindran, A.~R.}
\newblock \bibinfo{title}{Platelet ordering policies at hospitals using stochastic integer programming model and heuristic approaches to reduce wastage}.
\newblock \emph{\bibinfo{journal}{Comput. Ind. Eng.}} \textbf{\bibinfo{volume}{110}}, \bibinfo{pages}{151--164} (\bibinfo{year}{2017}).

\bibitem{Lundberg2017a}
\bibinfo{author}{Lundberg, S.} \& \bibinfo{author}{Lee, S.-I.}
\newblock \bibinfo{title}{A unified approach to interpreting model predictions}.
\newblock \emph{\bibinfo{journal}{Proceedings of the 31st {International} {Conference} on {Neural} {Information} {Processing} {Systems}}} \bibinfo{pages}{4765--4774} (\bibinfo{year}{Long Beach, CA, USA, 4–9 December 2017}).

\bibitem{kron_multicenter_2021}
\bibinfo{author}{Kron, A.} \emph{et~al.}
\newblock \bibinfo{title}{Multicenter observational study evaluating the impact of platelet transport bags on product wastage}.
\newblock \emph{\bibinfo{journal}{Transfusion}} \textbf{\bibinfo{volume}{61}}, \bibinfo{pages}{1383--1388} (\bibinfo{year}{2021}).

\bibitem{fenwick_value_2020}
\bibinfo{author}{Fenwick, E.} \emph{et~al.}
\newblock \bibinfo{title}{Value of {Information} {Analysis} for {Research} {Decisions}—{An} {Introduction}: {Report} 1 of the {ISPOR} {Value} of {Information} {Analysis} {Emerging} {Good} {Practices} {Task} {Force}}.
\newblock \emph{\bibinfo{journal}{Value Health}} \textbf{\bibinfo{volume}{23}}, \bibinfo{pages}{139--150} (\bibinfo{year}{2020}).

\bibitem{feizi_vertical_2023}
\bibinfo{author}{Feizi, A.}, \bibinfo{author}{Orfanoudaki, A.}, \bibinfo{author}{Saghafian, S.} \& \bibinfo{author}{Hudgson, N.}
\newblock \bibinfo{title}{Vertical {Patient} {Streaming} in {Emergency} {Departments}}.
\newblock \bibinfo{howpublished}{SSRN} (\bibinfo{year}{2023}).
\newblock \urlprefix\url{https://papers.ssrn.com/abstract=4465161}.

\bibitem{seneviratne_bridging_2020}
\bibinfo{author}{Seneviratne, M.~G.}, \bibinfo{author}{Shah, N.~H.} \& \bibinfo{author}{Chu, L.}
\newblock \bibinfo{title}{Bridging the implementation gap of machine learning in healthcare}.
\newblock \emph{\bibinfo{journal}{BMJ Innov.}} \textbf{\bibinfo{volume}{6}} (\bibinfo{year}{2020}).

\bibitem{aubron_platelet_2018}
\bibinfo{author}{Aubron, C.}, \bibinfo{author}{Flint, A. W.~J.}, \bibinfo{author}{Ozier, Y.} \& \bibinfo{author}{McQuilten, Z.}
\newblock \bibinfo{title}{Platelet storage duration and its clinical and transfusion outcomes: a systematic review}.
\newblock \emph{\bibinfo{journal}{Crit. Care}} \textbf{\bibinfo{volume}{22}}, \bibinfo{pages}{185} (\bibinfo{year}{2018}).

\bibitem{estcourt_guidelines_2017}
\bibinfo{author}{Estcourt, L.~J.} \emph{et~al.}
\newblock \bibinfo{title}{Guidelines for the use of platelet transfusions}.
\newblock \emph{\bibinfo{journal}{Br. J. Haematol.}} \textbf{\bibinfo{volume}{176}}, \bibinfo{pages}{365--394} (\bibinfo{year}{2017}).

\bibitem{national_institute_for_health_and_care_excellence_blood_2015}
\bibinfo{author}{{National Institute for Health and Care Excellence}}.
\newblock \bibinfo{title}{Blood transfusion ({NICE} guideline 24)} (\bibinfo{year}{2015}).
\newblock \urlprefix\url{https://www.nice.org.uk/guidance/ng24/chapter/Recommendations}.

\bibitem{Blake2010}
\bibinfo{author}{Blake, J.}, \bibinfo{author}{Heddle, N.}, \bibinfo{author}{Hardy, M.} \& \bibinfo{author}{Barty, R.}
\newblock \bibinfo{title}{Simplified platelet ordering using shortage and outdate targets}.
\newblock \emph{\bibinfo{journal}{Int. J. Healthc. Manag.}} \textbf{\bibinfo{volume}{1}}, \bibinfo{pages}{145--166} (\bibinfo{year}{2010}).

\bibitem{sekhar_effective_2016}
\bibinfo{author}{Sekhar, M.}, \bibinfo{author}{Clark, S.}, \bibinfo{author}{Atugonza, R.}, \bibinfo{author}{Li, A.} \& \bibinfo{author}{Chaudhry, Z.}
\newblock \bibinfo{title}{Effective implementation of a patient blood management programme for platelets}.
\newblock \emph{\bibinfo{journal}{Transfusion Med.}} \textbf{\bibinfo{volume}{26}}, \bibinfo{pages}{422--431} (\bibinfo{year}{2016}).

\bibitem{abbaspour_simple_2021}
\bibinfo{author}{Abbaspour, A.}, \bibinfo{author}{Jahan, A.} \& \bibinfo{author}{Rezaiee, M.}
\newblock \bibinfo{title}{A simple empirical model for blood platelet production and inventory management under uncertainty}.
\newblock \emph{\bibinfo{journal}{J. Amb. Intel. Hum. Comp.}} \textbf{\bibinfo{volume}{12}}, \bibinfo{pages}{1783--1799} (\bibinfo{year}{2021}).

\bibitem{mirjalili_2022_data-driven_paper}
\bibinfo{author}{Mirjalili, M.}, \bibinfo{author}{Abouee-Mehrizi, H.}, \bibinfo{author}{Barty, R.}, \bibinfo{author}{Heddle, N.~M.} \& \bibinfo{author}{Sarhangian, V.}
\newblock \bibinfo{title}{A data-driven approach to determine daily platelet order quantities at hospitals}.
\newblock \emph{\bibinfo{journal}{Transfusion}} \textbf{\bibinfo{volume}{62}}, \bibinfo{pages}{2048--2056} (\bibinfo{year}{2022}).

\bibitem{Chen2016}
\bibinfo{author}{Chen, T.} \& \bibinfo{author}{Guestrin, C.}
\newblock \bibinfo{title}{{XGBoost}: a scalable tree boosting system}.
\newblock \emph{\bibinfo{journal}{Proceedings of the 22nd {ACM} {SIGKDD} {International} {Conference} on {Knowledge} {Discovery} and {Data} {Mining}}} \bibinfo{pages}{785--794} (\bibinfo{year}{San Francisco, CA, USA, 13–17 August 2016}).

\bibitem{shwartz-ziv_tabular_2022}
\bibinfo{author}{Shwartz-Ziv, R.} \& \bibinfo{author}{Armon, A.}
\newblock \bibinfo{title}{Tabular data: {Deep} learning is not all you need}.
\newblock \emph{\bibinfo{journal}{Inform. Fusion}} \textbf{\bibinfo{volume}{81}}, \bibinfo{pages}{84--90} (\bibinfo{year}{2022}).

\bibitem{akiba_optuna_2019}
\bibinfo{author}{Akiba, T.}, \bibinfo{author}{Sano, S.}, \bibinfo{author}{Yanase, T.}, \bibinfo{author}{Ohta, T.} \& \bibinfo{author}{Koyama, M.}
\newblock \bibinfo{title}{Optuna: {A} {Next}-generation {Hyperparameter} {Optimization} {Framework}}.
\newblock \emph{\bibinfo{journal}{Proceedings of the 25th {ACM} {SIGKDD} {International} {Conference} on {Knowledge} {Discovery} and {Data} {Mining}}} \bibinfo{pages}{2623--2631} (\bibinfo{year}{Anchorage, AK, USA, August 4-8 2019}).

\bibitem{ma_use_2013}
\bibinfo{author}{Ma, H.}, \bibinfo{author}{Bandos, A.~I.}, \bibinfo{author}{Rockette, H.~E.} \& \bibinfo{author}{Gur, D.}
\newblock \bibinfo{title}{On use of partial area under the {ROC} curve for evaluation of diagnostic performance}.
\newblock \emph{\bibinfo{journal}{Stat. Med.}} \textbf{\bibinfo{volume}{32}}, \bibinfo{pages}{3449--3458} (\bibinfo{year}{2013}).

\bibitem{lange_gymnax_2022}
\bibinfo{author}{Lange, R.~T.}
\newblock \bibinfo{title}{gymnax: {A} {JAX}-based reinforcement learning environment library}, \bibinfo{version}{0.0.5}.
\newblock \bibinfo{howpublished}{GitHub} (\bibinfo{year}{2022}).
\newblock \urlprefix\url{http://github.com/RobertTLange/gymnax}.

\bibitem{van_dijk_blood_2009}
\bibinfo{author}{van Dijk, N.~V.}, \bibinfo{author}{Haijema, R.}, \bibinfo{author}{van~der Wal, J.} \& \bibinfo{author}{Sibinga, C.~S.}
\newblock \bibinfo{title}{Blood platelet production: a novel approach for practical optimization}.
\newblock \emph{\bibinfo{journal}{Transfusion}} \textbf{\bibinfo{volume}{49}}, \bibinfo{pages}{411--420} (\bibinfo{year}{2009}).

\bibitem{mirjalili_data-driven_2022}
\bibinfo{author}{Mirjalili, M.}
\newblock \emph{\bibinfo{title}{Data-driven modelling and control of hospital blood inventory}}.
\newblock Ph.D. thesis, \bibinfo{school}{University of Toronto}, \bibinfo{address}{Toronto, Canada} (\bibinfo{year}{2022}).
\newblock \urlprefix\url{https://tspace.library.utoronto.ca/bitstream/1807/124976/1/Mirjalili_Mahdi_202211_PhD_thesis.pdf}.

\bibitem{farrington_going_2023}
\bibinfo{author}{Farrington, J.}, \bibinfo{author}{Li, K.}, \bibinfo{author}{Wong, W.~K.} \& \bibinfo{author}{Utley, M.}
\newblock \bibinfo{title}{Going faster to see further: {GPU}-accelerated value iteration and simulation for perishable inventory control using {JAX}}.
\newblock \bibinfo{howpublished}{arXiv} (\bibinfo{year}{2023}).
\newblock \urlprefix\url{http://arxiv.org/abs/2303.10672}.

\bibitem{blake_use_2009}
\bibinfo{author}{Blake, J.~T.}
\newblock \bibinfo{title}{On the use of operational research for managing platelet inventory and ordering}.
\newblock \emph{\bibinfo{journal}{Transfusion}} \textbf{\bibinfo{volume}{49}}, \bibinfo{pages}{396--401} (\bibinfo{year}{2009}).

\bibitem{prioli_economic_2022}
\bibinfo{author}{Prioli, K.~M.} \emph{et~al.}
\newblock \bibinfo{title}{Economic implications of {FDA} platelet bacterial guidance compliance options: {Comparison} of single‐step strategies}.
\newblock \emph{\bibinfo{journal}{Transfusion}} \textbf{\bibinfo{volume}{62}}, \bibinfo{pages}{365--373} (\bibinfo{year}{2022}).

\bibitem{szczepiorkowski_platelet_2022}
\bibinfo{author}{Szczepiorkowski, Z.~M.} \& \bibinfo{author}{Pagano, M.~B.}
\newblock \bibinfo{title}{Platelet components and bacterial contamination: hospital perspective 2022}.
\newblock \emph{\bibinfo{journal}{Hematology}} \textbf{\bibinfo{volume}{2022}}, \bibinfo{pages}{430--436} (\bibinfo{year}{2022}).

\bibitem{van_der_meer_platelet_2011}
\bibinfo{author}{van~der Meer, P.~F.} \& \bibinfo{author}{Korte, D.~d.}
\newblock \bibinfo{title}{Platelet preservation: {Agitation} and containers}.
\newblock \emph{\bibinfo{journal}{Transfus. Apher. Sci.}} \textbf{\bibinfo{volume}{44}}, \bibinfo{pages}{297--304} (\bibinfo{year}{2011}).

\bibitem{sutton_reinforcement_2018}
\bibinfo{author}{Sutton, R.~S.} \& \bibinfo{author}{Barto, A.~G.}
\newblock \emph{\bibinfo{title}{Reinforcement Learning: An Introduction}} \bibinfo{edition}{2} edn (\bibinfo{publisher}{The MIT Press}, \bibinfo{year}{2018}).

\bibitem{sun_analyses_2012}
\bibinfo{author}{Sun, R.} \& \bibinfo{author}{Zhao, G.}
\newblock \bibinfo{title}{Analyses about efficiency of reinforcement learning to supply chain ordering management}.
\newblock \emph{\bibinfo{journal}{{IEEE} 10th {International} {Conference} on {Industrial} {Informatics}}} \bibinfo{pages}{124--127} (\bibinfo{year}{Beijing, China, July 25-27 2012}).

\bibitem{yadan_hydra_2019}
\bibinfo{author}{Yadan, O.}
\newblock \bibinfo{title}{Hydra - {A} framework for elegantly configuring complex applications}, \bibinfo{version}{1.3.2}.
\newblock \bibinfo{howpublished}{GitHub} (\bibinfo{year}{2019}).
\newblock \urlprefix\url{https://github.com/facebookresearch/hydra}.

\end{thebibliography}

\pagebreak

\begin{appendices}
\renewcommand{\appendixname}{Supplementary Note}
\titleformat{\section}[block] % Use the 'block' style
{\normalfont\Large\bfseries}
{} % Empty, as we'll incorporate the label into the title itself
{0em} % No separation space
{\appendixname~\thesection:~} % Prepend the title with "Supplementary Material A: "

\section{Platelet inventory management at UCLH} \label{app:uclh_processes}

The transfusion laboratory at UCLH maintains a stock of platelets and other blood products. All stock is ordered from NHS Blood and Transplant (NHSBT) using their Online Blood Ordering System (OBOS). The transfusion laboratory receives three routine deliveries each weekday. During the period under consideration there was one routine delivery on Saturday and none on Sunday. A routine Sunday delivery has since been introduced. The transfusion laboratory has a standing order for platelets to be delivered each day in the first routine delivery slot. In between these deliveries, ad hoc or emergency orders may be placed, which arrive within 1--2 hours. Ad hoc orders are generally made for patients with special transfusion requirements, or when there has been unusually high usage and additional units are expected to be required before the next routine delivery slot. Emergency orders are delivered more quickly than ad hoc orders, with a blue light service, but are made very rarely because UCLH is not a trauma centre. Units are logged on arrival into the laboratory information management system Bank Manager. Unallocated platelet units and platelet units ordered for patients with special transfusion requirements are stored in a platelet agitator located in the transfusion laboratory until they are required. Platelet agitators keep platelets at room temperature and gently agitate the bagged units to ensure the platelets are oxygenated \cite{van_der_meer_platelet_2011}.

When clinical staff wish to order platelets for a patient, they place a request with the transfusion laboratory via telephone or email, and the transfusion laboratory staff register the request in Bank Manager. This request includes the date and time at which the units are required, the delivery location, the ward on which the patient is staying, and the discipline of the consultant responsible for the patient's care. The transfusion laboratory staff identify a compatible unit using Bank Manager, or place an order with NHSBT via OBOS if required. The unit(s) are then delivered from the transfusion laboratory by courier to a local platelet agitator or to the patient's bedside for the required time. In general, the transfusion laboratory staff try to ensure that the allocated units are the same ABO blood group as the patient, and then select the oldest unit. If there are units in stock that expire at the end of the current day, they will be issued to meet the next request for which they are an acceptable match. The units allocated to each request are recorded in Bank Manager. Couriers move issued platelet units from the transfusion laboratory to agitators near a patient, or directly to a patient's bedside. Couriers and transfusion laboratory staff visit agitators outside the laboratory between 10:00 and 13:00 each day and return all remaining platelet units back to the transfusion laboratory.

\clearpage

\section{Additional simulation details}\label{app:sim_processes}

\subsection{Workflow as a Markov decision process}

An MDP formally describes a a sequential decision problem in terms of a set of states $s \in \mathbb{S}$, a set of actions $a \in \mathbb{A}$, a set of a rewards $r \in \mathbb{\Psi}$, a function defining the transition dynamics between states, and a discount factor $\gamma \in [0,1]$ which specifies the relative importance of immediate and future rewards \cite[Chapter~3]{sutton_reinforcement_2018}. The state of the environment $S_t$ comprises three components: the day of the week $\tau_t \in \{0, 1, ... 6\}$, the stock on hand $\underline{\text{X}}_t$ ordered by ascending age and the units that were issued on day $t-1$ but not transfused,  $\underline{\text{Z}}_t$ again ordered by ascending age.

\begin{equation}
\underline{\text{X}}_t = \left[X_{m, t}, X_{m-1, t}, ..., X_{1,t}\right]
\end{equation}

\begin{equation}
\underline{\text{Z}}_t = \left[Z_{m-1, t}, Z_{m-2, t}, ..., Z_{0,t}\right]
\end{equation}

We assume that $Z_t$ is not observable at the point the agent takes its decision, and therefore the agent observes $O_t = \left[\tau_t, \underline{\text{X}}_t\right]$. This is therefore a partially observable Markov decision Process (POMDP) \cite[Chapter~17]{sutton_reinforcement_2018}, in which the observation accurately captures part of the underlying state, and part of the underlying state is always unobservable. 

The action in the MDP, $A_t$ is the number of platelet units ordered at the start of day $t$. While the focus of this work is on issuing policies, this required finding plausible replenishment policies, and we used the MDP to fit the parameters of heuristic replenishment policies as described in Section \ref{sec:rep_policy} and Supplementary Note \ref{app:rep_policies}. Our heuristic replenishment policies use, at most, the day of the week and the total stock on hand at the start of the period as input.  

When order $A_t$ is placed, the units are assumed to be delivered immediately, a lead time $L$ of zero periods. The age profile of the units received is $\underline{\text{Y}}_t = \left[Y_{m, t}, Y_{m-1, t}, ..., Y_{1,t}\right]$. The remaining useful life of the units on arrival is modelled by a multinomial distribution with a number of trials equal to the order quantity $A_t$ and a number of events equal to the maximum useful life $m$. The parameters of the multinomial distribution may depend on the day of the week, $\tau_t$, and the parameters corresponding to weekday $\tau$ are represented as $\underline{\Delta}_{\tau}$. The transition dynamics of the system are defined implicitly by the simulation.

\subsection{Order of events}

There are six stages in our simulated workflow, as illustrated in Figure \ref{fig:sim_steps}. Some variables are updated at multiple stages, and we indicate the value of a variable at the end of stage $k$  using a superscript on the left hand side of the variable, e.g. $^k\underline{\text{X}}_t$ is the vector of units in stock at the end of stage $k$ on day $t$. Stage 0 is start of the day, before an order is placed and $^0\underline{\text{X}}_t \equiv \underline{\text{X}}_t$. The number of units for which an emergency order was placed due to a shortage is $E_t$ (${}^0E_t=0$) and the vector of units issued on day $t$ but not transfused is $\underline{\text{U}}_t$ ($^0\underline{\text{U}}_t = \underline{0}$). 

Demand is sampled independently for the morning and afternoon using parameters $\mu^{\text{am}}_{\tau} = \mu^{\text{pm}}_{\tau} = \frac{\mu_{\tau}}{2}$ for day of the week $\tau$. We use the function \textsc{meet\_demand} in the equations below to represent the process of issuing units and simulating the true and predicted labels required for our proposed issuing policy. To simulate predicted labels, we specified the sensitivity $\alpha$ and specificity $\beta$ of the predictive model we assumed to be available. We describe this approach in Section \ref{sec:issuing_policy} and provide pseudocode for \textsc{meet\_demand} in Algorithm \ref{alg:issuing}. 

\textbf{Stage 1}: At the start of each day $t$, the agent makes an observation $O_t$ comprising the current inventory in stock (split by remaining useful life) and the day of the week. The agent places a replenishment order, $A_t$, following a replenishment policy and the order, with age profile $\underline{\text{Y}}_t$, is assumed to arrive immediately.

\begin{align}
    \underline{\text{Y}}_t & \sim \text{Multinomial}(A_t, \underline{\Delta}_{\tau}) \\
    {}^1X_{m,t} &= Y_{m,t} \\
    {}^1X_{j,t} &= {}^0X_{j,t} + Y_{j,t} \quad \forall j, 1 \leq j \leq m-1
\end{align}

\textbf{Stage 2}: Total demand for the morning, $D_t^{\text{am}}$, is sampled from a Poisson distribution, and filled following the issuing policy. If there is a shortage, an emergency order is placed to fill the demand. For each request, we sample from a Bernoulli distribution with probability of success termed the return rate, $\rho$, to determine whether the unit will be returned or transfused to a patient. 

\begin{align}
    D_t^{\text{am}} &\sim \text{Poisson}(\mu_{\tau}^{\text{am}} ) \\
    {}^2\underline{\text{X}}_t, {}^2\underline{\text{U}}_{t}, {}^2E_t &= \textsc{meet\_demand}({}^1\underline{\text{X}}_t, {}^0\underline{\text{U}}_t, {}^0E_t, D_t^{\text{am}}, \alpha, \beta, \rho, \underline{\Delta}_{\tau})
\end{align}

\textbf{Stage 3}: At midday, the units that were issued on day $t-1$ but not transfused are returned to the blood bank. Returned units that were issued on their last day of remaining useful life will have expired. The number of returned units with remaining useful life $i$ that are not in a fit state to be reissued, $N_{i,t}$, is sampled from a Binomial distribution with a number of trials equal to the number of returned units with remaining useful life $i$ and a probability of success termed the slippage rate, $\phi$. Units that have not expired or been wasted due to slippage are now available to fill demand that arises during the rest of the day.

\begin{align}
    {}^3X_{m,t} &= {}^2X_{m,t} \\
    N_{i,t} &\sim \text{Binomial}(Z_{i,t}, \phi) \quad \forall i, 1 \leq i \leq m - 1 \\
    {}^3X_{i,t} &= {}^2X_{i,t} + Z_{i,t} - N_{i,t} \quad \forall i, 1 \leq i \leq m - 1
\end{align}

\textbf{Stage 4}: As at stage 2, total demand for the afternoon, $D_t^{\text{pm}}$ is sampled from a Poisson distribution, and filled following the issuing policy. If there is a shortage, an emergency order is placed to fill the demand. For each request, we sample from a Bernoulli distribution with probability of success termed the return rate, $\rho$, to determine whether the unit will be returned or transfused to a patient. 

\begin{align}
    D_t^{\text{pm}} &\sim \text{Poisson}(\mu_{\tau}^{\text{pm}} ) \\
    {}^4\underline{\text{X}}_t, {}^4\underline{\text{U}}_{t}, {}^4E_t &= \textsc{meet\_demand}({}^3\underline{\text{X}}_t, {}^2\underline{\text{U}}_t, {}^2E_t, D_t^{\text{pm}}, \alpha, \beta, \rho, \underline{\Delta}_{\tau})
\end{align}

\textbf{Stage 5}: The stock is aged one day, and any units that had a remaining useful life of one day at the start of day $t$ are assumed to expire. 

\begin{align}
    {}^5X_{i,t} &= {}^4X_{i+1,t} \quad \forall i,  1 \leq i < {m} \\
    {}^5X_{m,t} &= 0 \\
    W_t &= {}^4X_{1,t} + \sum^{m-1}_{i=1}N_{i,t}
\end{align}

\textbf{Stage 6:} The state is updated for the start of the next day and the reward is calculated. 

\begin{align}
\tau_{t+1} &= \tau_{t} + 1 \mod 7 \\
X_{i,t+1} &= {}^5X_{i,t} \quad \forall i,1 \leq i \leq m\\
Z_{i-1,t+1} &= {}^4U_{i,t} \quad \forall i,1\leq i \leq m
\end{align}

\begin{equation} \label{eq:reward}
R_{t+1} = -C_f \mathbbm{1}_{A_t>0} - C_vA_t - C_h \sum_{i=1}^{m-1} {}^5X_{i,t} - C_s {}^4E_t -  C_w \left[W_t + \frac{Z_{0,t}}{\gamma}\right]
\end{equation}

The reward function component related to units that were issued on their last day of life but were not transfused and therefore expired, $Z_{0,t}$ is divided by the discount factor $\gamma$ because it is registered one timestep later than it would have been if it had expired in stock. This ensures that, if considering the discounted return $G$, there is no difference between a unit expiring in stock or in a remote agitator waiting to be returned. 

In Algorithm \ref{alg:issuing} we set out how our YUPR issuing policy operates at stages 2 and 4 of the simulated workflow. 

\begin{algorithm}
\scriptsize
\caption{Meet demand for simulation stages 2 and 4}
\begin{algorithmic}[1]
\Function{issue\_yufo}{$\underline{\text{X}}$}
    \For{$i \gets 0 \text{ to } \text{length}($\underline{\text{X}}$)-1$}
        \If{$\underline{\text{X}}[i] > 0$}
            \State \Return $i$
        \EndIf
    \EndFor
\EndFunction

\Function{issue\_oufo}{$\underline{\text{X}}$}
    \For{$i \gets \text{length}(\underline{\text{X}})-1 \text{ downto } 0$}
        \If{$\underline{\text{X}}[i] > 0$}
            \State \Return $i$
        \EndIf
    \EndFor
\EndFunction

\Function{meet\_demand}{$\underline{\text{X}}, \underline{\text{U}}, E, D, \alpha, \beta, \rho, \underline{\Delta}$}
    \For{$d \gets 1 \text{ to } D$}
        \State $label \gets \text{Sample from Bernoulli}(\rho)$
        \If{$\text{sum}(\underline{\text{X}}) > 0$}
            \If{$label = 1$}
                \State $sample \gets \text{Sample from Uniform}(0, 1)$
                \If{$sample < \alpha$}
                    \State $prediction \gets 1$
                    \State $index \gets \text{Call } \textsc{issue\_yufo}(\underline{\text{X}})$
                \Else
                    \State $prediction \gets 0$
                    \State $index \gets \text{Call } \textsc{issue\_oufo}(\underline{\text{X}})$
                \EndIf
                \State $\underline{\text{U}}[index] \gets \underline{\text{U}}[index] + 1$
            \Else
                \State $sample \gets \text{Sample from Uniform}(0, 1)$
                \If{$sample > \beta$}
                    \State $prediction \gets 1$
                    \State $index \gets \text{Call } \textsc{issue\_yufo}(\underline{\text{X}})$
                \Else
                    \State $prediction \gets 0$
                    \State $index \gets \text{Call } \textsc{issue\_oufo}(\underline{\text{X}})$
                \EndIf
            \EndIf
            \State $\underline{\text{X}}[index] \gets \underline{\text{X}}[index] - 1$
        \Else
            \State $E \gets E + 1$
            \State $index \gets \text{Sample from Multinomial}(1, \underline{\Delta})$
            \If{$label = 1$}
                \State $\underline{\text{U}}[index] \gets \underline{\text{U}}[index] + 1$
            \EndIf
        \EndIf
    \EndFor
    \State \Return $\underline{\text{X}}, U, E$
\EndFunction
\end{algorithmic}
\label{alg:issuing}
\end{algorithm}

\clearpage
\section{Simulation inputs} \label{app:sim_inputs}

Four key simulation inputs were calculated based on information from Bank Manager, the laboratory information management system used by the UCLH transfusion laboratory. The mean daily demand, return rate, and slippage rate were calculated based on a common set of platelet requests: we queried all the requests for platelets where the units were required between 1 January 2015 and 31 December 2016 inclusive and to which a unit was allocated, and excluded any requests where the allocated unit was a neonatal unit or where the patient was a test patient used for system diagnostics.

\subsection{Mean daily demand}

\subsubsection{Total demand for units (UCLH Tx+r)}

The mean daily demand for each day of the week was calculated as the sum of units in the common set of platelet requests that were required on that day of the week divided by the total number of occurrences of that day of the week in the period. These values are set out in Table \ref{tab:demand_inc_returns}. 

\begin{table}[h!]
\caption{Mean daily demand per weekday including returns (UCLH Tx+r)}\label{tab:demand_inc_returns}
\begin{tabular*}{\textwidth}{@{\extracolsep\fill}lrrrrrrr}
\toprule%
 & Mon & Tue & Wed & Thu & Fri & Sat & Sun \\
\midrule
$\mu_{\tau}$ & 28.8 & 33.4 & 26.2 & 28.4 & 30.8 & 18.6 & 19.6 \\
\midrule
$\mu_{\tau}^{\text{am}}$ & 14.4 & 16.7 & 13.1 & 14.2 & 15.4 & 9.3 & 9.8 \\
$\mu_{\tau}^{\text{pm}}$ & 14.4 & 16.7 & 13.1 & 14.2 & 15.4 & 9.3 & 9.8 \\
\botrule
\end{tabular*}
\end{table}

\subsubsection{Demand for transfused units (UCLH Tx)}

To estimate the demand for units that were transfused, we multiplied the values obtained for the total demand (UCLH Tx+r) by $(1 - \rho)$. These values are set out in Table \ref{tab:demand_exc_returns}.

\begin{table}[h!]
\caption{Mean daily demand per weekday adjusted by the return rate to include only units that will be transfused (UCLH Tx)}\label{tab:demand_exc_returns}
\begin{tabular*}{\textwidth}{@{\extracolsep\fill}lrrrrrrr}
\toprule%
 & Mon & Tue & Wed & Thu & Fri & Sat & Sun \\
\midrule
$\mu_{\tau}$& 26.4 & 30.6 & 24.2 & 26.0 & 28.4 & 17.0 & 18.0 \\
\midrule
$\mu_{\tau}^{\text{am}}$ & 13.2 & 15.3 & 12.1 & 13.0 & 14.2 & 8.5 & 9.0 \\
$\mu_{\tau}^{\text{pm}}$ & 13.2 & 15.3 & 12.1 & 13.0 & 14.2 & 8.5 & 9.0 \\
\botrule
\end{tabular*}
\end{table}

\subsection{Return rate $(\rho)$}

The return rate was calculated as the proportion of requested units in the common set of requests where the final allocation status was not either ``Transfusion'' or ``Transfusion Assumed''. 

\subsection{Slippage rate ($\phi$)}

We took the total number of units where reissue was possible as requested units in the common set of requests where the final allocation status was not ``Transfusion'' or ``Transfusion Assumed'' and where the allocated unit was not due to expire on the day the units were required. We estimated the slippage rate as the proportion of these units that were not subsequently reissued to another request. The estimate is an upper limit because there may be some units included in the numerator that were in a sufficiently good state to be reissued but were not due to a lack of demand. This is a conservative approach to estimating the benefits of our proposed issuing policy because we expect it will have a greater advantage over an OUFO policy when the slippage rate is low. 
    
\subsection{Distribution of remaining useful life on arrival (UCLH)}
    
We queried all platelet units received between 1 January 2015 and 31 December 2016 inclusive, and excluded units that were either neonatal or not extended-life (extended life units, with a maximum shelf life of seven days from donation comprised approximately 99\% of the units received). The maximum useful life of extended-life units on arrival at the hospital blood bank is five days due to processing and testing required before distribution. To ensure the distributions represented the age of units available for routine morning deliveries we also excluded units not received within 90 minutes after the first routine delivery slot of the day. There were no routine deliveries on Sundays during the period of study, so we used the routine delivery time of Saturday as a proxy. The remaining useful life of a unit was calculated as the number of days between receipt of the unit and its expiry date (with 1 day corresponding to receipt on the day of expiry). For each day of the week, we calculated the units received on that day of the week with a remaining useful life of one to five days as a percentage of the total number of units received on that day of the week meeting the criteria. The same process was repeated for units received between 1 January 2017 and 31 December 2017 inclusive to calculate the distribution for 2017 used as part of the evaluation of the trained predictive model within the simulated workflow. The parameters for the multinomial distribution of remaining useful life on arrival for 2015--2016 is set out in Table \ref{tab:ul_on_arrival_uclh} and the corresponding figures for 2017 in Table \ref{tab:ul_on_arrival_uclh_2017}. For simplicity, the same distributions were used for both the routine morning deliveries and any emergency orders placed in the event of a shortage shortages. The distribution of remaining useful life on arrival may differ later in the day, but very few emergency orders were placed due to the high service levels achieved by the policies. 

In Table \ref{tab:ul_on_arrival_scen} we present the parameters for the multinomial distribution of remaining useful life on arrival used for Experiment 7, which were generated from a binomial distribution with a changing probability of success as described in Section \ref{methods:sim:sim_exp}. 

\begin{table}[h!]
\caption{Parameters for the multinomial distribution of remaining useful life on arrival from UCLH in 2015-2016}\label{tab:ul_on_arrival_uclh}
\begin{tabular*}{\textwidth}{@{\extracolsep\fill}lrrrrrrr}
\toprule%
Weekday & \multicolumn{7}{@{}c@{}}{Remaining useful life on arrival (days)} \\
\cmidrule{2-8}
& 5 & 4 & 3 & 2 & 1 \\
\midrule
Mon & 0.25 & 0.33 & 0.28 & 0.11 & 0.03 \\
Tue & 0.20 & 0.35 & 0.27 & 0.13 & 0.05 \\
Wed & 0.26 & 0.18 & 0.38 & 0.14 & 0.04 \\
Thu & 0.76 & 0.07 & 0.05 & 0.09 & 0.03 \\
Fri & 0.62 & 0.29 & 0.02 & 0.03 & 0.04 \\
Sat & 0.61 & 0.28 & 0.11 & 0.00 & 0.00 \\
Sun & 0.48 & 0.27 & 0.19 & 0.05 & 0.01 \\
\botrule
\end{tabular*}
\end{table}

\begin{table}[h!]
\caption{Parameters for the multinomial distribution of remaining useful life on arrival for UCLH in 2017}\label{tab:ul_on_arrival_uclh_2017}
\begin{tabular*}{\textwidth}{@{\extracolsep\fill}lrrrrrrr}
\toprule%
Weekday & \multicolumn{7}{@{}c@{}}{Remaining useful life on arrival (days)} \\
\cmidrule{2-8}
& 5 & 4 & 3 & 2 & 1 \\
\midrule
Mon & 0.31 & 0.32 & 0.23 & 0.12 & 0.02 \\
Tue & 0.18 & 0.48 & 0.21 & 0.10 & 0.03 \\
Wed & 0.25 & 0.19 & 0.38 & 0.15 & 0.03 \\
Thu & 0.87 & 0.02 & 0.03 & 0.07 & 0.01 \\
Fri & 0.71 & 0.24 & 0.02 & 0.01 & 0.01 \\
Sat & 0.62 & 0.28 & 0.10 & 0.00 & 0.00 \\
Sun & 0.48 & 0.28 & 0.18 & 0.06 & 0.00 \\
\botrule
\end{tabular*}
\end{table}

\begin{table}[h!]
\caption{Parameters for the multinomial distribution of remaining useful life on arrival for sensitivity analysis in Experiment 7}\label{tab:ul_on_arrival_scen}
\begin{tabular*}{\textwidth}{@{\extracolsep\fill}lrrrrrrr}
\toprule%
Probability & \multicolumn{7}{@{}c@{}}{Remaining useful life on arrival (days)} \\
of success &  \\
\cmidrule{2-8}
& 5 & 4 & 3 & 2 & 1 \\
\midrule
0.0 & 0.00  & 0.00  & 0.00  & 0.00  & 1.00  \\
0.1 & 0.00 & 0.00 & 0.05  & 0.29  & 0.66  \\
0.2 & 0.00 & 0.03 & 0.15 & 0.41 & 0.41  \\
0.3 & 0.01  & 0.08  & 0.26  & 0.41  & 0.24  \\
0.4 & 0.02  & 0.15  & 0.35  & 0.35  & 0.13 \\
0.5 & 0.06  & 0.25  & 0.38  & 0.25  & 0.06  \\
0.6 & 0.13 & 0.35  & 0.35  & 0.15  & 0.02  \\
0.7 & 0.24  & 0.41 & 0.26 & 0.08 & 0.01 \\
0.8 & 0.41 & 0.41 & 0.15 & 0.03 & 0.00 \\
0.9 & 0.66 & 0.29 & 0.05 & 0.00 & 0.00 \\
1.0 & 1.00 & 0.00 & 0.00 & 0.00 & 0.00 \\
\botrule
\end{tabular*}
\end{table}

\subsection{Simulation experiment settings}

In Table \ref{tab:sim_exp_summary} we summarise the settings for the simulation experiments based on the inputs described above. 

\begin{table}[h]
\footnotesize
\caption{Summary of settings for simulation experiments 1--7}\label{tab:sim_exp_summary}
\begin{tabular*}{\textwidth}{@{\extracolsep\fill}lllllll}
\toprule
Exp & Distribution &  $\rho$ & $\phi$ & Demand & Replenishment & Issuing      \\
& of remaining & & & distribution & policy & policy   \\
& useful life & & &  &  &  \\
& on arrival & & &  &  &  \\
\midrule 
1 & UCLH & 8\% & 7\% & UCLH Tx+r & Standing order & YUPR   \\
2 & UCLH & 8\% & 7\% & UCLH Tx+r & \texttt{(s,S)} & YUPR     \\
\midrule
3 & R\&R & 8\% & 7\% & UCLH Tx+r & Standing order & YUPR   \\
4 & R\&R & 8\% & 7\% & UCLH Tx+r & \texttt{(s,S)} & YUPR   \\
\midrule
5 & UCLH & \multicolumn{1}{c}{$\blacklozenge$} & 7\% & \multicolumn{1}{c}{$\lozenge$} & \texttt{(s,S)} & OUFO, YUPR-PPM   \\
6 & UCLH & 8\% & \multicolumn{1}{c}{$\blacklozenge$} & UCLH Tx+r & \texttt{(s,S)} & OUFO, YUPR-PPM   \\
7 & \quad {$\blacklozenge$} & 8\% & 7\% & UCLH Tx+r & \texttt{(s,S)} & OUFO, YUPR-PPM  \\
\botrule
\end{tabular*}
\footnotesize{The issuing policy YUPR denotes experiments that tested different combinations of predictive model sensitivity and specificity, including a setting equivalent to a standard OUFO issuing policy and a setting with a perfect predictive model. A black lozenge ($\blacklozenge$) denotes an input that was varied during an experiment, and a white lozenge ($\lozenge$) indicates an input that was varied during an experiment as a consequence of changing another input.}
\end{table}

\clearpage

\section{Replenishment policies} \label{app:rep_policies}

We considered two replenishment policies: a standing order policy and an $(\texttt{s}, \texttt{S})$ policy. The parameters for these heuristic policies were fit using simulation optimization, We evaluated each proposed set of parameters on 1,000 rollouts, each 365 days long following a warm-up period of 100 days. For each replenishment policy, the parameters(s) that achieved the highest mean return $G$ over the 1,000 rollouts for a given scenario were used for subsequent evaluation.

The standing order policy (Equation \ref{eq:sopolicy}) has a single parameter, $\texttt{Q}$, the number of units ordered each day. Under this policy, the same number of units are ordered on every day of the week, irrespective of the current stock. We used Optuna's grid sampler to evaluate all values of $\texttt{Q}$ between 0 and the maximum order quantity, $A_{\max} = 100$. The standing order policy that achieved the highest mean return $G$ over the 1,000 rollouts for a given scenario is characterised by the parameter $\texttt{Q}^{\text{best}}$, and was used for subsequent evaluation.

\begin{equation} \label{eq:sopolicy}
    A_t = \texttt{Q}
\end{equation}

Our $(\texttt{s}, \texttt{S})$ is formally an $(\texttt{R}, \texttt{s}, \texttt{S})$ policy with a fixed review period \texttt{R} equal to one day. The $(\texttt{s}, \texttt{S})$ policy has 14 parameters: an order-up-to level parameter $\texttt{S}$ and a reorder point parameter $\texttt{s}$ for each day of the week \cite{sun_analyses_2012}. The order-up-to-level for a given day of the week represents the maximum quantity that can be ordered on that day, if there are no units in stock. No units will be ordered when the number of units in stock is greater than the reorder point. The order quantity on day $t$, given that the day of the week is $\tau$ and the total current stock on hand is $X_t$ is:

\begin{equation} \label{eq:sspolicy}
A_t = \begin{cases}
\left[\texttt{S}^{\tau} - X_t\right]^+ &\text{if $X_t \leq \texttt{s}^{\tau}$} \\
0 &\text{if $X_t > \texttt{s}^{\tau}$}
\end{cases}
\end{equation}

\noindent where $(\texttt{s}^{\tau}, \texttt{S}^{\tau})$ is the pair of parameters for day of the week $\tau$.

We used Optuna's NSGAII sampler, a genetic algorithm, to suggest parameter combinations of $\texttt{s}^{\tau} \in \{0, 1, ..., \texttt{s}_{\max} = A_{\max}\} = 100$ and $\texttt{S}^{\tau} \in \{0, 1, ..., \texttt{S}_{\max} = A_{\max}\} = 100 \; \forall \tau \in \{0, 1, .., 6\}$. There is a hard constraint on the relative values of \texttt{s} and \texttt{S} for each weekday: $\texttt{s}^{\tau} < \texttt{S}^{\tau} \; \forall \tau \in \{0, 1, .., 6\}$. It is not possible to restrict the search space in Optuna based on relative values of parameters, and therefore we enforced this constraint by forcing the policy to order zero units on each day of the simulation if it was violated. For each generation of the genetic algorithm, we ran 50 proposed combinations of parameters in parallel and ranked them based on the mean return $G$. We terminated the search procedure when the best combination of parameters had not changed for a specified number generations, or after 200 generations had been completed. The $(\texttt{s}, \texttt{S})$  policy that achieved the highest mean return $G$ over the 1,000 rollouts for a given scenario when the search procedure was terminated is characterised by parameters $\left(\left(\texttt{s}^0, \texttt{S}^0\right), ..., \left(\texttt{s}^6, \texttt{S}^6\right)\right)_{\text{best}}$ and was used for subsequent evaluation.

For Experiments 1--4, we refit the parameters of the replenishment policy for each pair of values of sensitivity and specificity, using the best parameters identified for a policy that is equivalent to an OUFO policy  as the first proposed combination for each subsequent simulation optimization run. We terminated the parameter search when the best policy parameters had not changed for 10 generations, or after a maximum of 200 generations. 

For Experiments 5--7, for each different level of an input setting in each experiment we fit the parameters for the replenishment policy under an OUFO issuing policy first and used these as the starting point when fitting the parameters for both the next input setting under an OUFO issuing policy, and for the corresponding input setting under a YUPR-PPM issuing policy. We terminated the parameter search when the best policy parameters had not changed for 50 generations, or after a maximum of 200 generations. Note that this early-stopping limit is more relaxed than in Experiments 1--4 for computational reasons. 

\newpage

\section{Simulation experiment results} \label{app:sim_exp_results_all}

\subsection{Initial simulation experiment results}

We initially used the simulated workflow to estimate the performance of the system in terms of the KPIs using different combinations of replenishment and issuing policy assuming the distribution of remaining useful life on arrival observed at UCLH and that some requests did not result in transfusion (Experiments 1 and 2). We repeated these experiments assuming the distribution of remaining useful life on arrival reported by Rajendran and Ravindran \cite{rajendran_platelet_2017}, 
but keeping all other inputs the same (Experiments 3 and 4 respectively). 

Table \ref{tab:basic_sim} summarises the results of Experiments 1--4, in terms of the mean daily cost, service level and wastage over 10,000 simulated years. The standard deviation of each measure reflects the uncertainty in each scenario (due to uncertainty in demand, remaining useful life on arrival, slippage, and whether a requested unit is transfused or returned). See Section \ref{methods:sim:sim_exp} for details of the experiments and Table \ref{tab:sim_exp_summary} in Supplementary Note \ref{app:sim_inputs} for a summary of the different input settings. The costs reflect the balancing of different priorities of the problem and do not directly reflect monetary costs. We therefore focus our discussion of the results on the KPIs. We set out the details of the cost components in Section \ref{sec:results:workflow} and describe how the KPIs were calculated in Section \ref{methods:sim:sim_exp}. A lower bound on the costs for Experiments 1--4 is the expected daily cost for replenishment (16,066) based on the mean demand for units that will be transfused and assuming an order is placed every day. 

We estimated the KPIs for different levels of predictive model quality defined by sensitivity and specificity. Tables \ref{tab:basic_sim} and \ref{tab:basic_sim_pairwise} present the results at two key settings: a standard OUFO issuing policy (equivalent to a YUPR policy with $\text{sensitivity}=0.0$ and $\text{specificity}=1.0$) , and a YUPR policy with a perfect predictive model (YUPR-PPM; $\text{sensitivity}=$ $\text{specificity}=1.0$). Table \ref{tab:basic_sim_pairwise} sets out the paired-sampled differences in performance for combinations of replenishment and issuing policy over the 10,000 simulated years used for evaluation. 

There is a clear benefit to jointly improving the replenishment policy and the issuing policy: the best results are achieved when using an \texttt{(s,S)} replenishment policy and our YUPR issuing policy with a PPM. In Experiments 2 and 4 wastage due to time expiry is significantly reduced, with the remaining wastage due almost entirely to slippage. The KPIs for this pair of experiments demonstrate the particular effectiveness of our proposed policy at mitigating the negative effects of receiving stock with a shorter average remaining useful life on arrival. In Supplementary Note \ref{app:no_returns}, we present the results of using our workflow and scenario inputs under the assumption made in previous work: that all requested units are transfused. Wastage was low even with a simple standing order policy and was reduced to 0\% for both distributions of remaining useful life on arrival under an \texttt{(s,S)} replenishment policy and an OUFO issuing policy.

\begin{figure}[h] 
\centering
\includegraphics[width=1.0\textwidth]{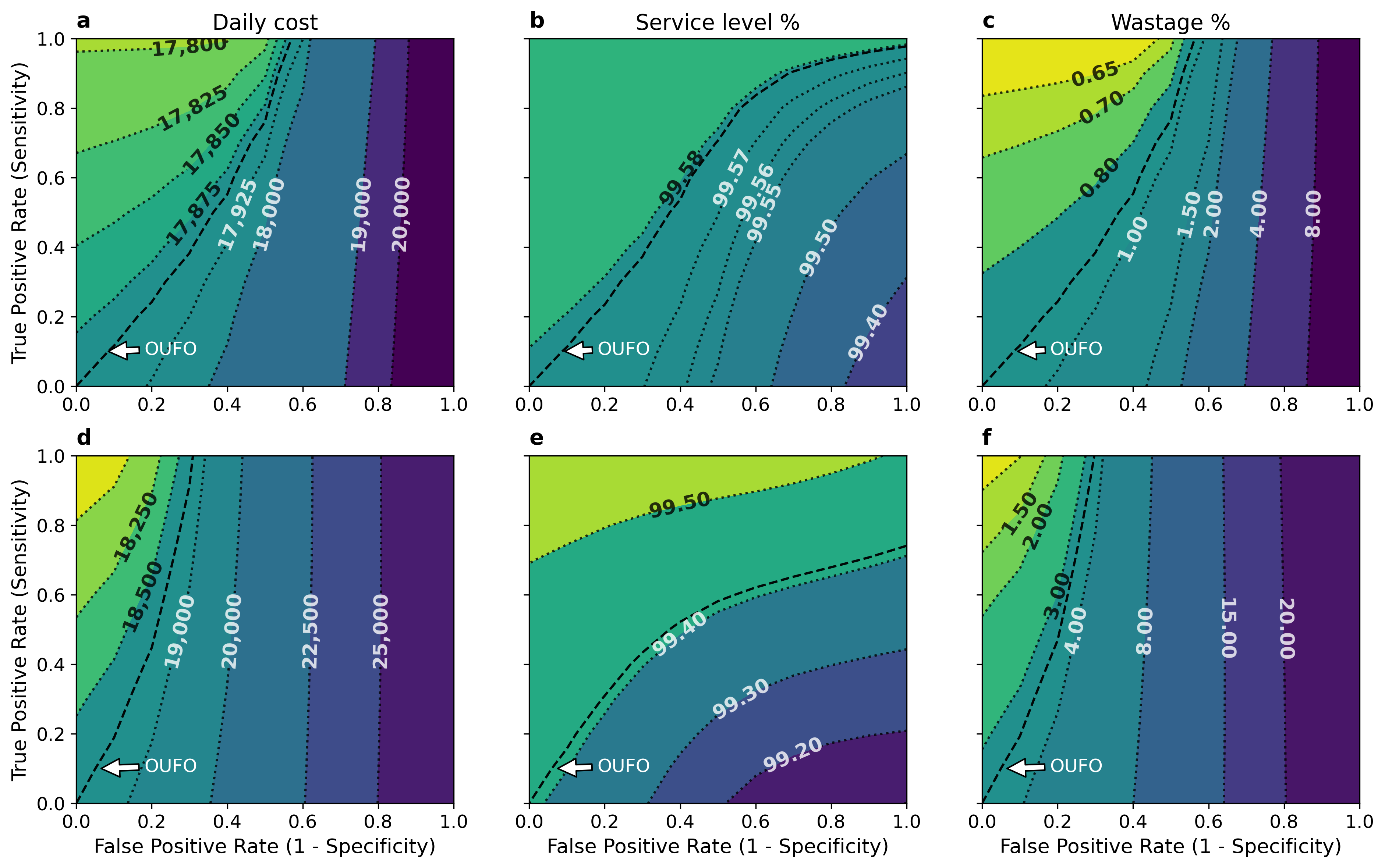} \\
\caption{\textbf{Contour plots illustrating the performance of our proposed issuing policy with different levels of predictive model quality}. The daily cost (\textbf{a}), service level (\textbf{b}) and wastage (\textbf{c}) assuming the distribution of remaining useful life on arrival observed at UCLH, and the corresponding metrics (\textbf{d,e,f}) when assuming the distribution of remaining useful life on arrival reported by Rajendran and Ravindran \cite{rajendran_platelet_2017}. These plots show that under both settings our proposed approach can reduce wastage and cost, with no reduction in service level, relative to an OUFO issuing policy. Lighter colours indicate better performance. A perfect predictive model, with sensitivity and specificity both equal to 1.0, would be in the top left corner of a subplot. The region above and to the left of the contour for an OUFO issuing policy comprises combinations of sensitivity and specificity required for our issuing policy to perform better than OUFO. Each plot contains a labelled contour showing the performance for a baseline OUFO issuing policy and the colour map for each plot is centred on that contour. Subplots \textbf{(a,b,c)} are based on results from Experiment 2 and subplots \textbf{(d,e,f)} are based on results from Experiment 4.} \label{fig:sim_contour}
\end{figure}

The full results from Experiments 2 and 4, for different levels of predictive model performance, are presented as contour plots in Figure \ref{fig:sim_contour}, with the contour plots for wastage also overlaid by the ROC of the trained predictive model in Figure \ref{fig:contour_test} in Section \ref{sec:results}. While our proposed issuing policy could cut wastage with no reduction in service level, a predictive model with a high false positive rate would lead to higher wastage and costs than the baseline OUFO policy. 

% Taken from 20230926_uclh_demand_aoa_results_table.ipynb (Exp 1- 4)
% Taken from 20230926_uclh_demand_rs_aoa_results_table.ipynb (Exp 5-8)
\begin{table}[h]
\footnotesize
\caption{Simulation results from Experiments 1--4.}\label{tab:basic_sim}
\begin{tabular*}{\textwidth}{@{\extracolsep\fill}llllrrrrrr}
\toprule%
Distribution  & Exp & Replenishment & Issuing & \multicolumn{2}{c}{Daily cost}  & \multicolumn{2}{c}{Service level \%} & \multicolumn{2}{c}{Wastage \%}  \\
of remaining & & policy & policy & & &  \\
useful life & & &  & & &  \\
on arrival & & &  & & &  \\
\midrule
\multirow{4}{*}{UCLH} & \multirow{2}{*}{1}& Standing order & OUFO & 20,344 & (500) & 97.5 & (0.8) & 1.0 & (0.2)\\
&  & Standing order & YUPR-PPM & 20,337 & (533) & 97.8 & (0.9) & 0.7 & (0.2) \\
\cmidrule(r){2-10}
& \multirow{2}{*}{2} & $(\texttt{s}, \texttt{S})$ & OUFO & 17,891 & (199) & 99.6 & (0.1) & 0.9 & (0.1) \\
&  & $(\texttt{s}, \texttt{S})$ & YUPR-PPM & 17,797 & (197) & 99.6 & (0.1) & 0.6 & (0.1)\\
\botrule
\multirow{4}{*}{R\&R} &  \multirow{2}{*}{3} & Standing order & OUFO & 20,992 & (425) & 97.5 & (0.6) & 4.8 & (0.5)\\
& & Standing order & YUPR-PPM & 20,145 & (301) & 98.8 & (0.4) & 3.5 & (0.7) \\
\cmidrule(r){2-10}
& \multirow{2}{*}{4} & $(\texttt{s}, \texttt{S})$ & OUFO & 18,719 & (215) & 99.4 & (0.2) & 3.4 & (0.2) \\
& & $(\texttt{s}, \texttt{S})$ & YUPR-PPM & 17,829 & (200) & 99.5 & (0.1) & 0.7 & (0.1)\\
\botrule
\end{tabular*}
\footnotetext{The mean (standard deviation) of the daily cost and KPIs were calculated from 10,000 evaluation rollouts, each 365 days long.}
\end{table}

\begin{table}
\caption{Paired-sample comparison of policy combinations from Experiments 1 \& 2 and Experiments 3 \& 4}\label{tab:basic_sim_pairwise}
\footnotesize
\begin{tabular*}{1.01\textwidth}{@{\extracolsep\fill}lllrrrr}
\toprule
 Metric & Replenishment \\
 & policy & & \multicolumn{2}{c}{Standing order} &  \multicolumn{2}{c}{\texttt{(s,S)}} \\
\cmidrule(lr){4-5} \cmidrule(lr){6-7}
& & Issuing  & & & & \\
 & & policy & \multicolumn{1}{c}{OUFO} & \multicolumn{1}{c}{PPM} & \multicolumn{1}{c}{OUFO} & \multicolumn{1}{c}{PPM} \\
\midrule
\\
\multicolumn{6}{l}{\textbf{Experiments 1 \& 2: UCLH distribution of remaining useful life on arrival}} \\
\\
\multirow{4}{*}{Daily cost} & \multirow{2}{*}{Standing order} & OUFO & & $-7 \pm 2$ & $-2,452 \pm 4$ & $-2,547 \pm 4$\\
& & YUPR-PPM & & & $-2,446 \pm 5$ & $-2,540 \pm 5$ \\
&\multirow{2}{*}{\texttt{(s,S)}} & OUFO & & & & $-94 \pm 0$\\
& & YUPR-PPM & \\
\cmidrule(r){2-7}
\multirow{4}{*}{Service level \%} & \multirow{2}{*}{Standing order} & OUFO & & $0.3 \pm 0.0$ & $2.1 \pm 0.0$ & $2.1 \pm 0.0$\\
& & YUPR-PPM & & & $1.8 \pm 0.0$ & $1.8 \pm 0.0$\\
&\multirow{2}{*}{\texttt{(s,S)}} & OUFO & & & & $0.0 \pm 0.0$ \\
& & YUPR-PPM & \\
\cmidrule(r){2-7}
\multirow{4}{*}{Wastage \%}& \multirow{2}{*}{Standing order} & OUFO & & $-0.3 \pm 0.0$ & $-0.1 \pm 0.0$ & $-0.4 \pm 0.0$\\
& & YUPR-PPM & & & $0.2 \pm 0.0$ & $-0.1 \pm 0.0$ \\
&\multirow{2}{*}{\texttt{(s,S)}} & OUFO & & & & $-0.3 \pm 0.0$ \\
& & YUPR-PPM & \\
\botrule
\\
\multicolumn{6}{l}{\textbf{Experiments 3 \& 4: R\&R distribution of remaining useful life on arrival}} \\
\\
\multirow{4}{*}{Daily cost} & \multirow{2}{*}{Standing order} & OUFO & & $-847 \pm 3$ & $-2,274 \pm 3$ & $-3,164 \pm 3$ \\
& & YUPR-PPM & & & $-1,426 \pm 3$ & $-2,316 \pm 3$ \\
&\multirow{2}{*}{\texttt{(s,S)}} & OUFO & & & & $-890 \pm 1$\\
& & YUPR-PPM & \\
\cmidrule(r){2-7}
\multirow{4}{*}{Service level \%} & \multirow{2}{*}{Standing order} & OUFO & & $1.3 \pm 0.0$ & $1.9 \pm 0.0$ & $2.0 \pm 0.0$ \\
& & YUPR-PPM & & & $0.6 \pm 0.0$ & $0.7 \pm 0.0$\ \\
&\multirow{2}{*}{\texttt{(s,S)}} & OUFO & & & & $0.1 \pm 0.0$ \\
& & YUPR-PPM & \\
\cmidrule(r){2-7}
\multirow{4}{*}{Wastage \%} & \multirow{2}{*}{Standing order} & OUFO & & $-1.3 \pm 0.0$ & $-1.4 \pm 0.0$ & $-4.1 \pm 0.0$ \\
& & YUPR-PPM & & & $-0.2 \pm 0.0$ & $-2.8 \pm 0.0$ \\
&\multirow{2}{*}{\texttt{(s,S)}} & OUFO & & & & $-2.7 \pm 0.0$ \\
& & YUPR-PPM & \\
\botrule
\end{tabular*}
\footnotesize{Each value represents the mean of the paired-sample differences ($\pm$ the standard error of the mean paired-sample difference) of the daily cost or KPI between the policy combination on the column and the policy combination on the row, over the 10,000 evaluation rollouts. Negative values for daily cost and wastage, and positive values for service level, indicate the the policy combination on the column performed better than the policy combination on the row.}
\end{table}

\subsection{Additional sensitivity analyses} \label{app:scenario_analysis}

\begin{figure}[h]
\centering
\includegraphics[width=1.0\textwidth]{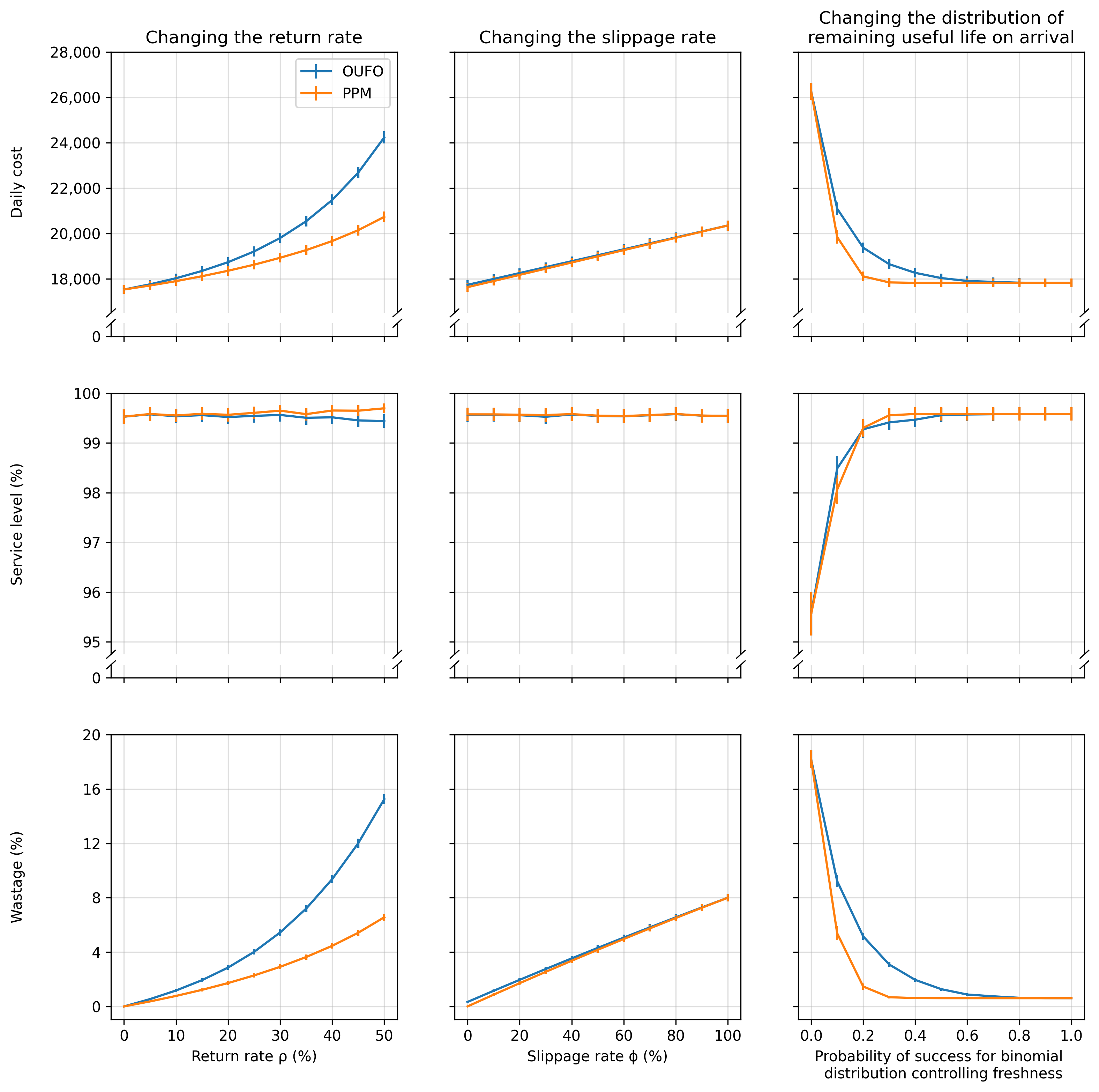}
\caption{\textbf{Impact of changing simulation input parameters on the the daily cost, service level and wastage when using an OUFO issuing policy and our YUPR issuing policy with a PPM.} Values are the mean of each metric over the 10,000 evaluation rollouts, and error bars are the standard deviation of the metric over 10,000 evaluation rollouts.}\label{fig:scenario_plot}
\end{figure}

\begin{figure}[h]
\centering
\includegraphics[width=1.0\textwidth]{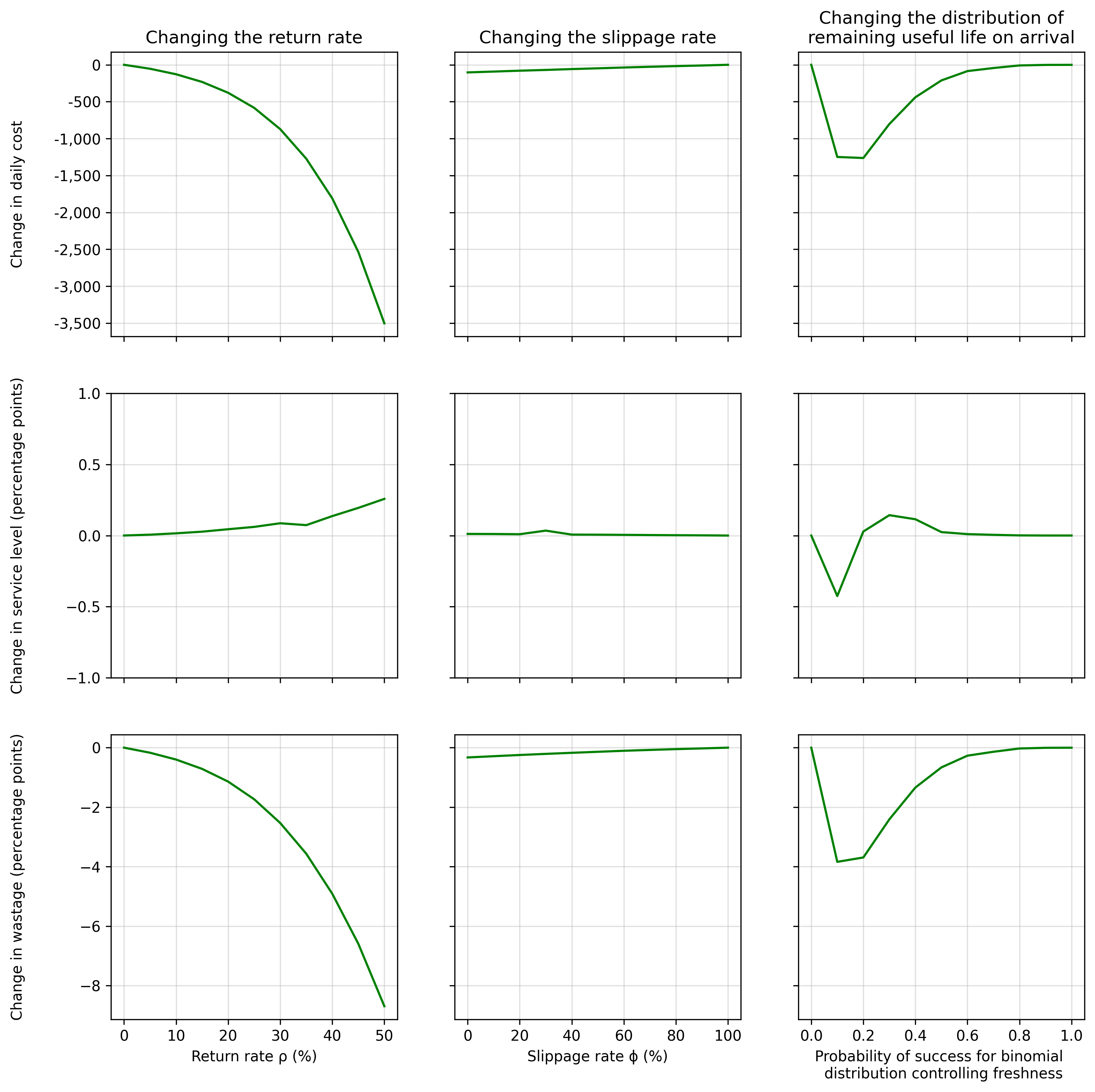}
\caption{\textbf{The paired-sample differences in daily cost, service level and wastage between an OUFO issuing policy and our YUPR issuing policy with a PPM.} Reductions in daily cost and wastage, and increases in service level, indicate cases where our issuing policy with a PPM performs better than an OUFO issuing policy. Values are the mean of the paired-sampled difference in the metrics over 10,000 evaluation rollouts, and error bars are the standard error of the mean paired-sample difference. The error bars are not visible at the scale selected to show the change in cost, level and wastage respectively.}\label{fig:pairwise_scenario_plot}
\end{figure}

In Figures \ref{fig:scenario_plot} and \ref{fig:pairwise_scenario_plot} we present the results from the sensitivity analyses: Experiments 5, 6 and 7. Figure 
\ref{fig:scenario_plot} illustrates the mean values of the daily cost, service level and wastage over the evaluation rollouts for an OUFO issuing policy and our proposed issuing policy with a PPM. Figure \ref{fig:pairwise_scenario_plot} illustrates the mean paired-sampled differences in these metrics between the two issuing policies over the evaluation rollouts. 

The YUPR issuing policy with a PPM had an increased beneficial impact  as the return rate increased, as the slippage rate decreased, and as the average age of stock at arrival increased. As expected, the YUPR policy with PPM performed the same as OUFO in cases when all requests result in transfusion (return rate $\rho$ = 0\%) or when units that are not transfused cannot be reissued (due to a high slippage rate, or because all of the stock expired on the day it arrived). The YUPR policy with a PPM also performed similarly to OUFO when most of the stock arrived fresh, because in this instance there is often sufficient time for a returned unit to be reissued even under an OUFO policy. The analysis shows the importance of achieving a low slippage rate: at a slippage rate $\phi=100\%$ the wastage under both issuing policies is 8\% (equal to  the return rate), but at a slippage rate of $\phi=0\%$ this is reduced to 0.3\% under an OUFO issuing policy and 0.0\% under our YUPR issuing policy with a PPM.

\clearpage

\section{Simulation experiments with no returns} \label{app:no_returns}

Previous work has made the assumption that all requested platelet units are transfused, and focused on optimizing replenishment policies. We conducted Experiments I--IV to determine how our system performed under that assumption. Experiments I--IV are similar to Experiments 1--4 respectively, but with no returns. We therefore set $\rho = 0$ and $\phi=0$ and only use an OUFO issuing policy because we anticipate no benefit to using a predictive model or our YUPR policy when there are no returns. We estimated the demand for units that would be transfused (UCLH Tx) by multiplying the total demand (UCLH Tx+r) by a factor of $(1-\rho)=0.92$, using the estimated return rate from UCLH during 2015-2016. The expected number of transfused units is therefore the same in Experiments 1--4 and I--IV, and is set out in Table \ref{tab:demand_exc_returns} in Supplementary Note \ref{app:sim_inputs}.

In Table \ref{tab:sim_exp_summary_no_returns} we summarise the key settings of Experiments I--IV and in Table \ref{tab:basic_sim_no_returns} we report the results of Experiments I--IV in terms of the mean daily cost, service level and wastage over 10,000 simulated years. 

\begin{table}[h]
\footnotesize
\caption{Summary of settings for Experiments I--IV}\label{tab:sim_exp_summary_no_returns}
\begin{tabular*}{\textwidth}{@{\extracolsep\fill}lllllll}
\toprule%
Exp & Distribution &  $\rho$ & $\phi$ & Demand & Replenishment & Issuing      \\
& of remaining & & & distribution & policy & policy   \\
& useful life & & &  &  &  \\
& on arrival & & &  &  &  \\
\midrule 
I & UCLH & 0\% & 0\% & UCLH Tx   & Standing order & OUFO \\
II & UCLH & 0\% & 0\% & UCLH Tx & \texttt{(s,S)} & OUFO   \\
\midrule
III & R\&R & 0\% & 0\% & UCLH Tx & Standing order & OUFO   \\
IV & R\&R & 0\% & 0\% & UCLH Tx & \texttt{(s,S)} & OUFO   \\
\botrule
\end{tabular*}
\end{table}

\begin{table}[h]
\footnotesize
\caption{Simulation results from Experiments I--IV.}\label{tab:basic_sim_no_returns}
\begin{tabular*}{\textwidth}{@{\extracolsep\fill}llllrrrrrr}
\toprule%
Distribution  & Exp & Replenishment & Issuing & \multicolumn{2}{c}{Daily cost}  & \multicolumn{2}{c}{Service level \%} & \multicolumn{2}{c}{Wastage \%}  \\
of remaining & & policy & policy & & &  \\
useful life & & &  & & &  \\
on arrival & & &  & & &  \\
\midrule
\multirow{2}{*}{UCLH} & I & Standing order & OUFO & 20,202 & (601) & 98.0 & (0.9) & 0.1 & (0.2) \\
& II & $(\texttt{s}, \texttt{S})$ & OUFO & 17,541 & (194) & 99.6 & (0.1) & 0.0 & (0.0) \\
\botrule
\multirow{2}{*}{R\&R} & III & Standing order & OUFO & 19,831 & (466) & 97.1 & (0.7) & 1.0 & (0.4) \\
& IV  & $(\texttt{s}, \texttt{S})$ & OUFO & 17,556 & (200) & 99.5 & (0.2) & 0.0 & (0.0) \\
\botrule
\end{tabular*}
\footnotetext{The mean (standard deviation) of the daily cost and KPIs were calculated from 10,000 evaluation rollouts, each 365 days long.}
\end{table}

The results of Experiments I--IV show that, if all requested units are transfused, then a simple $\texttt{(s,S)}$ replenishment policy is sufficient to achieve no wastage and a very high ($>99.5\%$) service level. 

\clearpage

\section{Training and test set request exclusion}\label{app:ml_exclusion}

In Table \ref{tab:ml_exclusion} we set out the how the training and test sets were extracted from the requests recorded in Bank Manager, the laboratory information management system used in the UCLH transfusion laboratory.

\begin{table}[h]
\caption{Requests used for predictive model}\label{tab:ml_exclusion}
\begin{tabular*}{\textwidth}{@{\extracolsep\fill}lrr}
\toprule%
& 2015 - 2016 & 2017\\
\midrule
Total requests & 18,917 & 9,793 \\
Less: no units assigned & (656) & (282) \\
Less: neonatal unit assigned & (236) & (152) \\
Less: request for test patient & -- & -- \\
Less: request required more than 30 minutes before registered & (5) & (6) \\
Less: requests removed so that all requests have 30 day lookback window & (723) & -- \\
\cmidrule{2-3}
& 17,297 & 9,353 \\

\botrule
\end{tabular*}
\end{table}

\clearpage

\section{Input features for machine learning model}\label{app:ml_features}

In Table \ref{tab:feature_summary} we summarise the input features for the predictive model. In Table \ref{tab:num_bin_features} we provide the proportion of missing values and the mean and standard deviation for the numeric and binary input features. In Table \ref{tab:cat_features} we provide the proportion of missing values and the three most frequent categories for the categorical features. 

\begin{sidewaystable}
\caption{Summary of input features}\label{tab:feature_summary}
\begin{tabular*}{\textheight}{llp{9cm}}
\toprule%
Feature name & Type (units) & Description \\
\midrule
age & Integer (years) & Difference between year of request and year of birth \\
male & Binary & Patient sex is male \\
request\_registered\_hour & Integer (0-23) & Hour of day request was registered \\
request\_registered\_weekday & Integer (0-6) & Day of the week request was registered \\
request\_required\_hour & Integer (0-23) & Hour of day requested units are required \\
request\_required\_weekday & Integer (0-6) & Day of week requested units are required \\
registered\_required\_diff\_hours & Float (hours) & Time between request being registered and the units being required \\
request\_priority & Binary & Request has priority flag \\
requested\_size & Integer (count) & Number of units requested \\
plt\_count\_value & Float ($10^9 \text{L}^{-1}$) & Value of the most recent platelet count available at the prediction point (up to 7 days ago) \\
plt\_result\_time\_diff\_hours & Float (hours) & Time between the result of the most recent platelet count and the prediction point \\
plt\_specimen\_time\_diff\_hours & Float (hours) & Time between the recorded collection time of the specimen used for the most recent platelet count and the prediction point \\
num\_units\_allocated\_last\_30\_days & Integer (count) & Number of units allocated to previous requests for this patient in the last 30 days \\
last\_transfused\_unit\_required\_time\_diff\_hours & Float (hours) & Time between when the most recent request for which a unit was transfused was required and the prediction point \\
proportion\_of\_requests\_last\_30\_days\_not\_all\_transfused & Float (proportion) & Proportion of requests required in the last 30 days where not all units were transfused \\
num\_requested\_units\_required\_after\_last\_plt\_sample & Integer (count) & Number of units allocated to requests required after the specimen time for the most recent platelet count \\
num\_units\_transfused\_since\_last\_plt\_sample & Integer (count) & Number of units recorded as transfused since the specimen time for the most recent platelet count \\
hospital & String & Hospital site the patient at which there patient is located \\
discipline & String & Discipline of the consultant responsible for the patient \\
ward\_name & String & Ward on which the patient is located \\
ward\_type & String & Type of ward on which the patient is located \\
plt\_count\_request\_location & String & Location from which the request for the most recent platelet count (within 7 days) was made \\
required\_location & String & Location to which the platelets are to be delivered \\
\botrule
\end{tabular*}
\end{sidewaystable}

\begin{sidewaystable}
\caption{Detail on numeric and binary features}\label{tab:num_bin_features}
\begin{tabular*}{\textheight}{lrrrrrrr}
\toprule%
& \multicolumn{3}{@{}c@{}}{2015 - 2016} & & \multicolumn{3}{@{}c@{}}{2017} \\\cmidrule{2-4}\cmidrule{6-8}%
Feature name & Missing \%  & \multicolumn{2}{@{}c@{}}{Mean (SD)} & & Missing \%  & \multicolumn{2}{@{}c@{}}{Mean (SD)}\\
\midrule
age & 0.43 & 52.32 & (18.79) & & 0.01 & 50.12 & (19.06) \\
male & 0.41 & 0.56 & & & 0.04 & 0.61 & \\
request\_registered\_hour & --  & 12.21 & (4.31) & & -- & 12.18 & (4.53) \\
request\_registered\_weekday & --  & 2.72 & (1.93) & & -- & 2.77 & (1.93) \\
request\_required\_hour & --   & 12.88 & (4.26) & & -- & 12.82 & (4.37) \\
request\_required\_weekday & --  & 2.75 & (1.93) & & -- & 2.79 & (1.94) \\
registered\_required\_diff\_hours & --  & 4.25 & (17.09) & & -- & 5.00 & (21.09)  \\
request\_priority & -- & 0.05 & & & -- & 0.03 & \\
requested\_size & --   & 1.08 & (0.28) & & -- & 1.05 & (0.22)\\
plt\_count\_value & 2.43 & 27.20 & (45.34) & & 3.27 & 24.74 & (35.36) \\
plt\_result\_time\_diff\_hours & 2.43 & 17.78 & (29.70) & & 3.27 & 18.06 & (28.14) \\
plt\_specimen\_time\_diff\_hours & 2.43 & 21.50 & (29.89) & & 3.27 & 22.64 & (29.13) \\
num\_units\_allocated\_last\_30\_days & --  & 8.02 & (8.76) & & -- & 7.20 & (7.90) \\
last\_transfused\_unit\_required\_time\_diff\_hours & 15.54 & 72.58 & (99.07) & & 16.14 & 77.66 & (106.42)  \\
proportion\_of\_requests\_last\_30\_days\_not\_all\_transfused & 15.10 & 0.07 & (0.14) & & 15.72 & 0.09 & (0.16)  \\
num\_requested\_units\_required\_after\_last\_plt\_sample & --  & 0.19 & (0.49) & & -- & 0.22 & (0.53) \\
num\_units\_transfused\_since\_last\_plt\_sample & -- & 0.22 & (0.27) & & -- & 0.24 & (0.55) \\
\botrule
\end{tabular*}
\end{sidewaystable}

\begin{sidewaystable}
\caption{Detail on categorical features}\label{tab:cat_features}
\begin{tabular*}{\textheight}{lrlrcllr}
\toprule%
& \multicolumn{3}{c}{2015 - 2016} & \multicolumn{3}{c}{2017} \\
\cmidrule{2-4} \cmidrule{6-8}
Feature name & Missing \%  & \multicolumn{2}{c}{Top 3 categories} & &  Missing \% & \multicolumn{2}{c}{Top 3 categories} \\
\midrule
hospital & -- & New UCLH & 68.6\% & & -- & New UCLH & 74.0\% \\
& & UCH Macmillan Cancer Centre & 14.5\% & & & UCH Macmillan Cancer Centre & 13.2\% \\
& &Harley Street on 15 & 12.2\% & & & Harley Street on 15 & 9.2\% \\
\midrule
discipline & 0.67 & Haematology (Clinical) & 73.0\% & & 0.84 & Haematology (Clinical) & 74.4\% \\
& & Paed Clinical Haematology & 5.3\% & & & Paed Clinical Haematology & 5.8\% \\
& & Medical Oncology & 3.2\% & & & Bone Marrow Transplant & 2.9\% \\
\midrule
ward\_name & -- & UCH Tower 16th Fl. S & 14.3\% & & -- & Oncology Tower 16th Fl. N & 11.8\% \\
& & UCH Tower 13th Fl. N & 11.0\% & & & UCH Tower 16th Fl. S & 11.4\% \\
& & Harley Street on 15 & 10.6\% & & & UCH Tower 13th Fl. N & 10.0\% \\
\midrule
ward\_type & 0.33 & Inpatient & 58.9\% & & 0.62 & Inpatient & 63.6\% \\
& & Day Case & 14.6\% & & & Day Case & 13.2\% \\
& & Outpatient & 12.7\% & & & Outpatient & 11.9\% \\
\midrule
plt\_count\_request\_location & 2.43 & T16S & 14.8\% & & 3.27 & T16N & 12.0\%\\
& & T13N & 12.0\% & & & T16S & 12.0\% \\
& & HS15 & 11.1\% & & & T13N & 10.0\% \\
\midrule
required\_location & -- & T16 Platelet Agitator & 63.9\% & & -- & T16 Platelet Agitator & 68.4\% \\
& & Non-Fridge Location & 26.0\% & & & MCC Platelet Agitator & 18.8\% \\
& & MCC Platelet Agitator & 7.8\% & & & Non-Fridge Location & 10.6\% \\
\botrule
\end{tabular*}
\end{sidewaystable}

\clearpage

\section{Hyperparameter tuning and threshold selection for machine learning model}\label{app:ml_hps}

In Table \ref{tab:ml_hps} we set out the search ranges for the hyperparameters of the ML pipeline we constructed using the Python library scikit-learn. The search ranges are expressed using the syntax of the Python library hydra \cite{yadan_hydra_2019}, which was used for configuration. We used the default parameter values for the \texttt{XGBClassifier} class from the Python library xgboost v1.7.5 for the hyperparameters not specified in the table 

The categorical features discipline, ward\_name and plt\_count\_request\_location were all encoded using the \texttt{OneHotEncoder} class from the Python library scikit-learn. The parameter min\_frequency is the minimum frequency required for a category to be encoded separately (categories below this threshold are collectively assigned to a separate `infrequent' class). 

\begin{table}[h]
\footnotesize
\caption{Hyperparameter search ranges and final values}\label{tab:ml_hps}
\begin{tabular*}{\textwidth}{@{\extracolsep\fill}lllr}
\toprule%
& Parameter & Search range & Final\\
&  &  & value\\
\midrule
Preprocessing & discipline - min\_frequency & \texttt{interval(0.01, 0.5)} & 0.15 \\
& ward\_name - min\_frequency & \texttt{interval(0.01, 0.5)} & 0.22\\
& plt\_count\_request\_location - min\_frequency & \texttt{interval(0.01, 0.5)} & 0.02 \\
\midrule
XGBoost & gamma & \texttt{tag(log, interval(0.1, 100))} & 38.11\\
& learning\_rate & \texttt{tag(log, interval(0.01, 0.3))} & 0.11 \\
& max\_depth & \texttt{int(interval(2,20))} & 6\textcolor{white}{.00} \\
& min\_child\_weight & \texttt{tag(log, interval(0.1, 100))} & 0.16 \\
& n\_estimators & \texttt{choice(50, 100, 200, 400, 800)} & 100\textcolor{white}{.00} \\
& reg\_alpha & \texttt{tag(log, interval(0.01, 100))} & 0.01 \\
& reg\_lambda & \texttt{tag(log, interval(0.01, 100))} & 0.80\\
& scale\_pos\_weight & \texttt{tag(log, interval(0.01, 100))} & 3.15 \\
& subsample & \texttt{interval(0.1, 1.0)} & 0.83 \\
\botrule
\end{tabular*}
\end{table}

\begin{figure}[h] 
\centering
\includegraphics[width=1.0\textwidth]{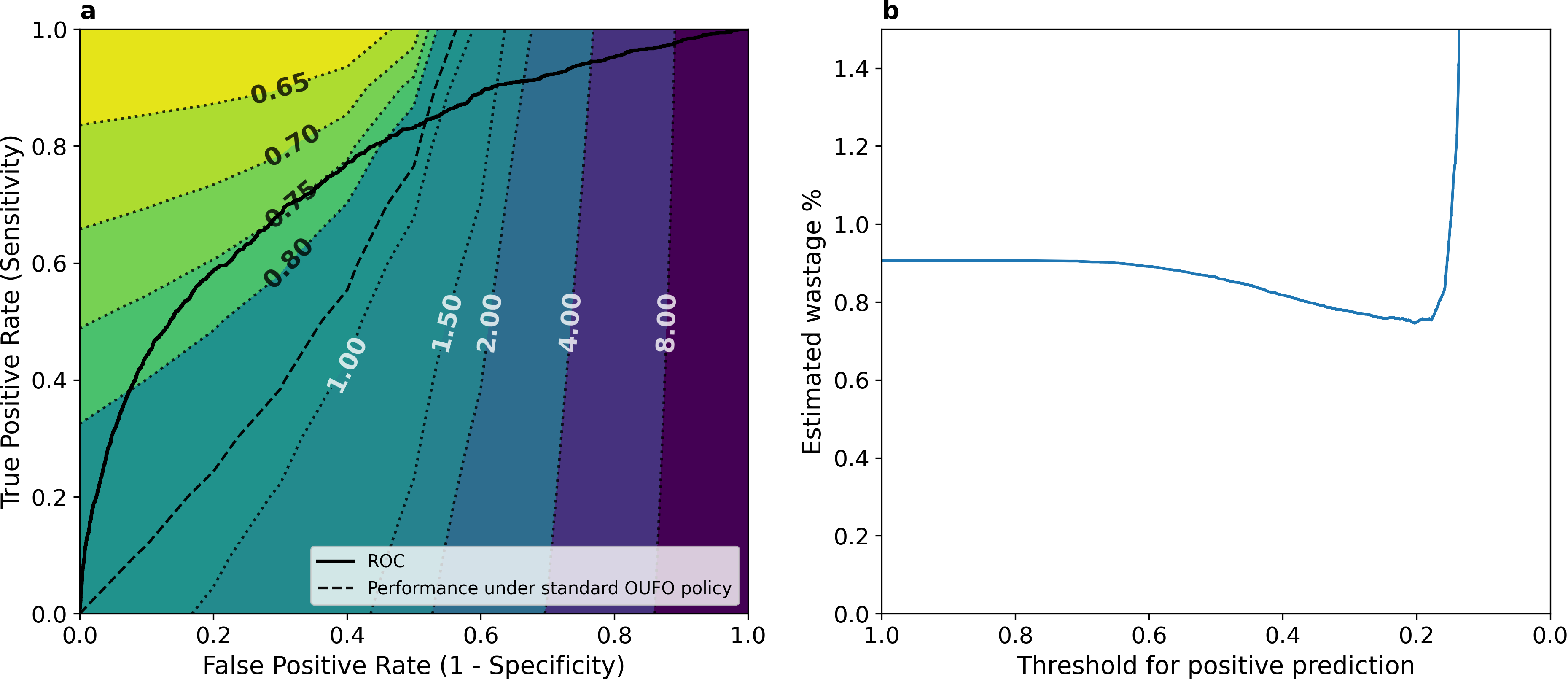}
\caption{\textbf{Determining the threshold for the predictive model using the simulation results from Experiment 2}. \textbf{(a)} The ROC on the training set overlaid on a contour plot of the wastage results from Experiment 2 and \textbf{(b)} the estimated wastage based on the results from Experiment 2 for different positive prediction thresholds on the training set (points along the ROC).} \label{fig:threshold_uclh}
\end{figure}

\begin{figure}[h] 
\centering
\includegraphics[width=1.0\textwidth]{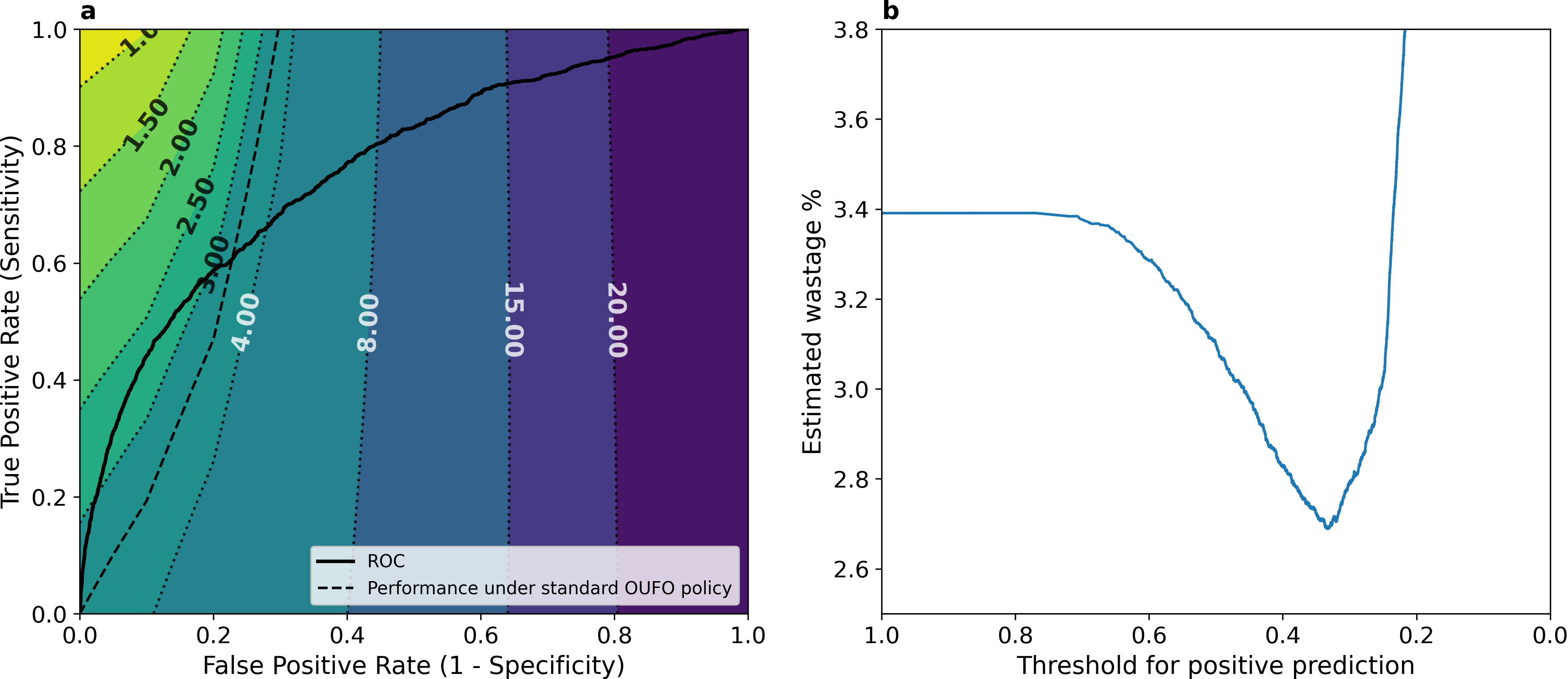}
\caption{\textbf{Determining the threshold for the predictive model using the simulation results from Experiment 4}. \textbf{(a)} The ROC curve on the training set overlaid on a contour plot of the wastage results from Experiment 4 and \textbf{(b)} the estimated wastage based on the results from Experiment 4 for different positive prediction thresholds on the training set (points along the ROC).}  \label{fig:threshold_rs}
\end{figure}

In order to apply the trained model within the simulated workflow, we needed to determine a threshold for distinguishing between positive and negative predictions. In Figure \ref{fig:threshold_uclh}a we plot the ROC curve for the training set over a contour plot of the wastage from Experiment 2 and in Figure \ref{fig:threshold_uclh}b we plot the estimated wastage against the threshold used to determine a positive prediction. In Figure \ref{fig:threshold_rs} presents the same information using the wastage results from Experiment 4, in which the distribution of remaining useful life on arrival was taken from previous work by Rajendran and Ravindran \cite{rajendran_platelet_2017}. 

\clearpage

\section{Evaluating the predictive model with real demand in simulated workflow}\label{app:real_demand_eval}

An alternative method for evaluating the performance of the predictive model is to incorporate the real demand and associated predictions into a simulated workflow, as described in Section \ref{subsec:methods:pred}. Using real requests for 2017 and the predictions of our trained models on these to inform YUPR issuing decisions, we estimate wastage of 1.1\% and a service level of 99.4\% (compared to 1.2\% and 99.4\% respectively using an OUFO issuing policy). Under the distribution of remaining useful life on arrival reported by Rajendran and Ravindran \cite{rajendran_platelet_2017} our YUPR issuing policy with a trained predictive model gave wastage of 4.3\% and a service level of 99.1\% (compared to 5.0\% and 99.1\% respectively using an OUFO issuing policy). In both cases, wastage was reduced compared to the OUFO baseline with no reduction in service level. Although the wastage is higher than expected from the simulation results plotted in Figure 1, this is consistent with the higher proportion of requests that did not result in transfusion, and higher wastage, observed at UCLH in 2017 compared to the preceding years. 

\clearpage

\section{Notation table}\label{app:notation}

In Table \ref{tab_notation} we set out a summary of the notation used in this work. 

\begin{table}[h]
\caption{Notation}\label{tab_notation}
\footnotesize
\begin{tabular*}{\textwidth}{@{\extracolsep\fill}lll}
\toprule%
Markov decision process & $S_t$ & State at the start of day $t$\\
& $\mathbb{S}$ & Set of possible states \\
&$O_t$ & Observation at the start of day $t$\\
&$A_t$ & Action taken at the start of day $t$ after observing $O_t$\\
& $\mathbb{A}$ & Set of possible actions \\
&$R_t$ & Reward received when observation $O_t$ is observed\\
& $\mathbb{\Psi}$ & Set of possible rewards \\
&$\gamma$ & Discount factor\\
&$G_t$ & Return, the discounted sum of rewards received after \\
& & taking action $A_t$. Unrelated to platelet units being \\
& & returned after issue. \\
\midrule
Reward function & $C_h$ & Holding cost per unit stock \\
& $C_s$ & Shortage cost per unit of unmet demand \\
& $C_w$ & Wastage cost per unit\\
& $C_v$ & Variable order cost per unit\\
& $C_f$ & Fixed order cost\\
\midrule
Workflow inputs & $\rho$ & Return rate, the probability that a requested unit will be\\
& & returned instead of transfused \\
& $\phi$ & Slippage rate, probability that returned unit that has not  \\
& & expired cannot be reissued \\
& $\underline{\Delta}_{\tau}$ & Vector of parameters for the multinomial distribution \\ 
& & of remaining useful life on arrival on day of week $\tau$\\
& $\mu_{\tau}$ & Mean of the Poisson distribution for demand on  \\
& & day of week $\tau$ \\
& $m$ & Maximum useful life on arrival \\
& $L$ & Lead time  \\
& $A_{\text{max}}$ & Maximum daily order quantity \\
\midrule
Inventory control & $\underline{\text{X}}_t$ & Vector of stock on hand at the start of day $t$, ordered by \\
& & ascending age. Component of state $S_t$ \\
& $X_t$ & Total stock on hand at the start of day $t$ \\
& $\underline{\text{Z}}_t$ & Vector of stock issued on day $t-1$ and not transfused, \\  
& & ordered by ascending age. Component of state $S_t$ \\
& $\underline{\text{Y}}_t$ & Vector of stock received to fill order $A_t$, ordered by \\
& & ascending age\\
& $\tau_t$ & Day of the week for day $t$. Component of state $S_t$ \\
& $D^\text{am}_t$ & Demand on the morning of day $t$ \\
& $D^\text{pm}_t$ & Demand on the afternoon of day $t$ \\
& $E_t$ & Number of units for which an emergency order was placed \\
& & on day $t$ due to a shortage \\
& $W_t$ & Number of units wasted on day $t$ due to expiry while in \\ 
& & stock or slippage \\
& $N_{i,t}$ & Number of units with remaining useful life $i$ days on day $t$ \\
& & that are lost to slippage \\
& $\underline{\text{U}}_t$ & Vector of units issued on day $t$ and not transfused, \\
& & ordered by ascending age \\
\midrule
Simulated machine learning model & $\alpha$ & Sensitivity \\
& $\beta$ & Specificity \\
\botrule
\end{tabular*}
\end{table}




\end{appendices}


\end{document}